\documentclass[a4paper,11pt,fleqn]{article}

\usepackage[ansinew]{inputenc}
\usepackage{hyperref}
\usepackage[mathscr]{eucal}
\usepackage{graphicx}
\usepackage[font=footnotesize,labelsep=period,justification=justified,figureposition=bottom,tableposition=top]{caption}
\usepackage{subcaption}
\usepackage{multirow}
\usepackage{amsmath,amssymb,amsthm}
\usepackage{comment}
\usepackage{caption}
\usepackage{subcaption}
\usepackage{xcolor}

\allowdisplaybreaks

\hypersetup{colorlinks, linkcolor=blue, citecolor=blue, urlcolor=blue}

\newcommand{\p}{\partial}

\newcommand{\uref}{u_{\text{ref}}}

\newcommand{\todo}[1][\null]{\ensuremath{\clubsuit}}

\newcommand{\noprint}[1]{}
\newcommand{\checked}[1][\null]{\ensuremath{\boldsymbol{\surd}}}

{\theoremstyle{definition}

\newtheorem{example}{Example}
\newtheorem{remark}{Remark}
\newtheorem*{remark*}{Remark}
\newtheorem*{notation*}{Notation}
}

\begin{document}

\par\noindent {\LARGE\bf
Improving physics-informed DeepONets \\with hard constraints 
\par}

\vspace{4mm}\par\noindent {\large
R\"udiger Brecht$^{\dag}$, Dmytro R.\ Popovych$^{\ddag,\S}$,  Alex Bihlo$^\ddag$ and Roman O.\ Popovych$^{\sharp,\S}$
\par}

\vspace{4mm}\par\noindent{\it
$^\dag$Department of Mathematics, University of Hamburg, Hamburg, Germany
}

\vspace{2mm}\par\noindent{\it
$^\ddag$ Department of Mathematics and Statistics, Memorial University of Newfoundland,\\
$\phantom{^\ddag}$~St.\ John's (NL) A1C 5S7, Canada
}

\vspace{2mm}\par\noindent{\it
$^\sharp$
Mathematical Institute in Opava, Silesian University in Opava, Na Rybnicku 626/1, 746 01 
$\phantom{^\sharp}$~Opava, Czech Republic
}

\vspace{2mm}\par\noindent{\it
$^\S$
Institute of Mathematics of NAS of Ukraine, 3 Tereshchenkivska Str., 01024 Kyiv, Ukraine
}

\vspace{2mm}\par\noindent {\it
\textup{E-mails:} ruediger.brecht@uni-hamburg.de, dpopovych@mun.ca,\\\phantom{\textup{E-mails:}} abihlo@mun.ca, rop@imath.kiev.ua
}\par

\vspace{12mm}\par\noindent\hspace*{10mm}\parbox{140mm}{\small
Current physics-informed (standard or deep operator) neural networks still rely on accurately learning the initial and/or boundary conditions of the system of differential equations they are solving. In contrast, standard numerical methods involve such conditions in computations without needing to learn them. In this study, we propose to improve current physics-informed deep learning strategies such that initial and/or boundary conditions do not need to be learned and are represented exactly in the predicted solution. Moreover, this method guarantees that when a deep operator network is applied multiple times to time-step a solution of an initial value problem, the resulting function is at least continuous. 
\par}\vspace{7mm}

\section{Introduction}

Recent years have seen tremendous interest in solving differential equations with neural networks. Originally introduced in~\cite{laga98a,laga00a} and popularized through~\cite{rais18a}, in which it is referred to as the method of \textit{physics-informed neural networks}, it has become popular throughout the mathematical sciences, with applications to astronomy~\cite{mart22a}, biomedical engineering~\cite{liu20a}, geophysics~\cite{song21a} and meteorology~\cite{bihl22a, brec23b}, just to name a few.

While the underlying method is conceptually straightforward to implement, several failure modes of the original method have been identified in the past~\cite{bihl23a,kris21a,wang22b,wang21a, wang22a}, along with some mitigation strategies. 

Setting these training difficulties aside, another fundamental shortcoming of physics-informed neural networks is that they require extensive training, to the point where they are seldom computationally competitive compared to standard numerical methods, see~\cite{brec23b} for an example related to weather prediction. The main issue is that each changing of initial and/or boundary conditions for a system of differential equations requires retraining of the neural network solution approximator, which is a costly endeavour, especially when accurate solutions are required.

One strategy to overcome this issue is to not learn the solution of a differential equation itself, but rather the solution operator. This idea, relying on the universal approximation theorem for operators~\cite{chen95a}, was proposed in~\cite{lu21a}, where it is referred to as \textit{physics-informed deep operator approach}, or \textit{physics-informed DeepONet}. The main idea is that a physics-informed DeepONet, which in the following we will simply refer to as DeepONet, accepts both the initial/boundary data and the independent variables of the differential equation to be solved, and learns to approximate the solution operator of that differential equation for any given initial/boundary conditions. Once trained, a DeepONet can thus simply be evaluated for different initial/boundary conditions, with no other cost than inference of the neural network. A notable advantage of this approach is that it allows time-stepping for time-dependent differential equations~\cite{wang23a}, which overcomes the failure mode of physics-informed neural networks that tend to converge to a trivial solution when trained for long time intervals. Thus, rather than solving a differential equation over the entire time interval with a single neural network, one trains a DeepONet over a shorter time interval and then repeatedly evaluates the DeepONet, using as new initial condition the final solution from the previous iteration step. This makes numerical methods based on DeepONets similar to more traditional numerical techniques such as Runge--Kutta methods or linear multistep methods. 
In \cite{li21a}, the physics-informed DeepONet approach was extended to parameterized families of partial differential equations.

While DeepONets have the potential to make neural network based differential equations solver computationally more efficient, they still suffer from some of the same training difficulties as standard physics-informed neural networks. 
One of the most challenging training difficulties to address is that the minimization problem to be solved for both physics-informed neural networks and DeepONets is typically formulated based on a non-convex, composite multi-task loss function. Here, the multiple tasks refer to simultaneously minimizing loss components associated with the differential equation and all initial and boundary conditions, respectively. 

However, already the original work by~\cite{laga98a} has shown that the minimization problem to be solved can be simplified by including these initial and/or boundary conditions as hard constraints to the neural network. 
There are a number of papers exploring the capacity of neural networks of various architectures with hard constraints 
to solve initial and boundary value problems for differential equations.
In particular, the hard-constraint approach was preferred in~\cite{mose23a}
in the course of developing finite-basis physics-informed neural networks, 
which are based on scalable domain decomposition inspired by the classical finite element method. 
Ideas on implementing periodic boundary conditions as a hard constraint for neural networks were presented 
for the first time in~\cite{dong21a} and then in~\cite{lu21b} for the one-dimensional case. 
A proper version of this technique in the form of a coordinate-transform layer 
and its realization for a sphere were developed in~\cite{bihl22a}.
Now using coordinate-transform layers has become a common practice when considering physics-informed neural networks with periodic boundary conditions. 
The residual networks when applied to approximating evolution operators, also known as flow maps~\cite{chen23a}, bear similarity to 
the simplest hard-constrained physics-informed networks for evolution equations.

Still, the prevalent way of solving both physics-informed neural networks and DeepONets relies on including these initial and/or boundary conditions in the loss function as a soft constraint. The aim of this paper is to present a comprehensive comparison between the two approaches, and to show that the hard-constraint approach is vastly superior to the soft-constraint approach for DeepONets.

The further organization of this paper is as follows. In Section~\ref{sec:Theory} we review the use of neural networks for solving differential equations, using physics-informed neural networks and DeepONets. Here we focus on both the soft and hard constraint formulations. In Section~\ref{sec:Results} we show some comparison for soft- and hard-constrained DeepONets solving some important benchmark problems such as the damped pendulum, the Lorenz--1963 model, the one-dimensional Poisson equation, the (1+1)-dimensional linear wave equation and the Korteweg--de Vries (KdV) equation. The final Section~\ref{sec:Conclusion} contains a summary of our work together with some possible further research avenues.

\section{Solving differential equations with neural networks}\label{sec:Theory}

In this section, we review the ideas of physics-informed neural networks and deep operator networks. 
For this, we consider the general form of a boundary value problem for a system of differential equation 
defined over the domain $\Omega$,
\begin{subequations}\label{eq:GenBVP}
\begin{gather}\label{eq:GenSystemOfDEs}
\Delta^l\big(\mathbf x,\mathbf u_{(n)}\big)=0,\quad l=1,\dots, L,\qquad \mathbf x\in\Omega,
\\[1ex]\label{eq:GenBCs}
\mathsf B ^{l_{\rm b}}\big(\mathbf x,\mathbf u_{(n_{\rm b})}\big)=0,\quad 
l_{\rm b}=1,\dots, L_{\rm b},\qquad \mathbf x\in\p\Omega.
\end{gather}
\end{subequations}
Here 
$\mathbf x=(x_1,\dots,x_d)$ is the tuple of independent variables, $\Omega\subset\mathbb R^d$, 
$\mathbf u=\mathbf u(\mathbf x)=(u^1,\dots, u^q)$ denotes the tuple of dependent variables, 
and $\mathbf u_{(n)}$ denotes the tuple of derivatives of the dependent variables~$\mathbf u$ 
with respect to the independent variables~$\mathbf x$ of order not greater than $n$.
$\mathsf B =(\mathsf B^1,\dots,\mathsf B^{L_{\rm b}})$ denotes the boundary value operator, $L_{\rm b}\in\mathbb N:=\{1,2,\dots\}$. 

There are various more specific versions of the general form~\eqref{eq:GenBVP} 
for particular classes of boundary value problems. 
Thus, if an independent variable is singled out in the entire tuple of independent variables as the time, 
we can re-denote the number of independent variables as $d+1$, 
start numbering them from zero and assume $x_0$ as the time variable, $x_0:=t$. 
Then the general boundary value problem~\eqref{eq:GenBVP} is converted into  
an initial--boundary value problem over the spatio-temporal domain $\Omega=[t_0,t_{\rm f}]\times\Omega'$,
where $[t_0,t_{\rm f}]$ and~$\Omega'\subset\mathbb R^d$ are the interval and the domain 
for the time variable~$t$ and the spatial variables $\mathbf x=(x_1,\dots,x_d)$, respectively,
and the general boundary conditions~\eqref{eq:GenBCs} split into 
initial conditions at $t=t_0$ and (spatial) boundary conditions at the boundary~$\p\Omega'$ of~$\Omega'$,  
\begin{align}\label{eq:GenIBVP}
\begin{split}
&\Delta^l\big(t,\mathbf x,\mathbf u_{(n)}\big)=0,\quad l=1,\dots, L,\qquad t\in[t_0,t_{\rm f}],\ \mathbf x\in\Omega',
\\[1ex]
&\mathsf I^{l_{\rm i}}\big(\mathbf x,\mathbf u_{(n_{\rm i})}\big|_{t=t_0}^{}\big)=0,\quad 
l_{\rm i}=1,\dots, L_{\rm i},\qquad \mathbf x\in\Omega',
\\[1ex]
&\mathsf B_{\rm s}^{l_{\rm sb}}\big(t,\mathbf x,\mathbf u_{(n_{\rm sb})}\big)=0,\quad 
l_{\rm sb}=1,\dots, L_{\rm sb},\qquad t\in[t_0,t_{\rm f}],\ \mathbf x\in\p\Omega'.
\end{split}
\end{align}
Here, $\mathsf I=(\mathsf I^1,\dots,\mathsf I^{L_{\rm i}})$ is the initial value operator 
and $\mathsf B_{\rm s}=(\mathsf B_{\rm s}^1,\dots,\mathsf B_{\rm s}^{L_{\rm sb}})$ denotes the (spatial) boundary value operator. 

In particular, for a system of evolution equations, one has $L=q$, 
$\Delta^l=u^l_t-\tilde\Delta^l(t,\mathbf x,\mathbf u_{(\mathbf x,n)})$ 
and $\mathsf I = \mathbf u|_{t=t_0}^{}-\mathbf f(\mathbf x)$, 
where $\mathbf u_{(\mathbf x,n)}$ denotes the tuple of derivatives of the dependent variables 
with respect to the spatial independent variables of order not greater than~$n$, 
$u^l_t:=\p u^l/\p t$, and
$\mathbf f(\mathbf x)=\big(f^1(\mathbf x),\dots,f^q(\mathbf x)\big)$ is a fixed vector function of~$\mathbf x$ 
with the domain~$\Omega'$.
For a system of differential equations that involves at most second-order differentiation with respect to~$t$ 
and can be solved with respect to~$\p^2u^l/\p t^2$, one also has $L=q$, and
the initial value operator is 
$\mathsf I = \big(\mathbf u|_{t=t_0}^{}-\mathbf f(\mathbf x),\mathbf u_t|_{t=t_0}^{}-\tilde{\mathbf f}(\mathbf x)\big)$, 
where $\tilde{\mathbf f}(\mathbf x)=\big(\tilde f^1(\mathbf x),\dots,\tilde f^q(\mathbf x)\big)$ 
is one more fixed vector function of~$\mathbf x$ with the domain~$\Omega'$. 
If $n=2$, a possible choice for boundary conditions is given by Dirichlet boundary conditions, where 
$\mathsf B_{\rm s} = \mathbf u-\mathbf g(t,\mathbf x)$
for a fixed vector function $\mathbf g(t,\mathbf x)=\big(g^1(t,\mathbf x),\dots, g^q(t,\mathbf x)\big)$ 
with the domain $[t_0,t_{\rm f}]\times\p\Omega'$.
Of course, there are many other kinds of boundary conditions that are relevant for applications 
(Neumann boundary conditions, mixed boundary conditions, periodic boundary conditions, etc.).
In the case of systems of ordinary differential equations, where $d=0$, 
one can consider purely initial value problems without boundary conditions 
or, conversely, boundary value problems of the general form, including multipoint boundary value problems.

\subsection{Physics-informed neural networks}

Let $\mathbf u^{\boldsymbol\theta}(\mathbf x)$ be the output of a deep neural network 
with inputs $\mathbf x$ and trainable parameters (weights) $\boldsymbol\theta$ 
that approximates the solution $\mathbf u$ of the boundary value problem~\eqref{eq:GenBVP}. 
The network can be trained over the neural network parameters~$\boldsymbol\theta$ using the composite loss function
\begin{gather}\label{eq:compositeLossFunction}
\mathcal L(\boldsymbol\theta) = \lambda_{\Delta}\mathcal L_\Delta(\boldsymbol\theta) + \lambda_{\rm b}\mathcal L_{\rm b}(\boldsymbol\theta)
\end{gather}
with
\begin{align*}
\mathcal L_\Delta(\boldsymbol\theta) &=\frac1{N_\Delta}\sum_{i=1}^{N_\Delta}\sum_{l=1}^L
\big|\Delta^l\big(\mathbf x^i_\Delta,\mathbf u^{\boldsymbol\theta}_{(n)}(\mathbf x^i_\Delta)\big)\big|^2,
\\[1.5ex]
\mathcal L_{\rm b}(\boldsymbol\theta) &= \frac1{N_{\rm b}}\sum_{i=1}^{N_{\rm b}}\sum_{l_{\rm b}=1}^{L_{\rm b}}
\big|\mathsf B^{l_{\rm b}}\big(\mathbf x^i_{\rm b},\mathbf u_{(n_{\rm b})}^{\boldsymbol\theta}(\mathbf x^i_{\rm b})\big)\big|^2
\end{align*}
being the equation loss and the boundary value loss, respectively. 
The parameters $\lambda_{\Delta},\lambda_{\rm b}\in\mathbb R_{>0}$ are the loss weights, 
which also can be assigned as trainable. 
The above losses are standard mean-squared errors computed at the sets of collocation points 
$\big\{\mathbf x_\Delta^i\big\}_{i=1}^{N_\Delta}\subset\Omega$ and 
$\big\{\mathbf x_{\rm b}^i\big\}_{i=1}^{N_{\rm b}}\subset\p\Omega$
for the system $\Delta$ and for the boundary conditions, respectively. 
The approximation of the solution~$\mathbf u$ of the boundary value problem~\eqref{eq:GenBVP} 
by the output $\mathbf u^{\boldsymbol\theta}(\mathbf x)$ of the deep neural network
is ensured by the simultaneous minimization of both losses.
In practice, when using mini-batch gradient descent,  
the collocation points are partitioned into subsets called batches. 
Every iteration over the entire set of collocation points, called an epoch, 
consists of multiple gradient descent steps, each carried out using the loss computed for the respective batch.

An important advantage of neural networks over standard numerical approximations lies 
in computing derivatives using automatic differentiation~\cite{bayd18a}. 
This avoids discretization errors in the computed derivatives, with the only approximation error stemming from finite-precision arithmetic.

When considering a particular boundary value problem,
the structure of the loss function~\eqref{eq:compositeLossFunction} for the general case, 
including the number of its components,  
should often be modified by taking into account specific features of the problem under consideration. 
Thus, for the initial--boundary value problem~\eqref{eq:GenIBVP}, 
it is natural to split the common boundary value loss into two parts, 
the initial value loss~$\mathcal L_{\rm i}(\boldsymbol\theta)$ and 
the spatial boundary value loss~$\mathcal L_{\rm sb}(\boldsymbol\theta)$.
In other words, the loss function in this case can be represented as 
\begin{gather*}
\mathcal L(\boldsymbol\theta) = 
 \lambda_{\Delta}\mathcal L_\Delta(\boldsymbol\theta) 
+\lambda_{\rm i}\mathcal L_{\rm i}(\boldsymbol\theta)
+\lambda_{\rm sb}\mathcal L_{\rm sb}(\boldsymbol\theta),
\end{gather*}
with
\begin{align*}
\mathcal L_\Delta(\boldsymbol\theta) ={}&\frac1{N_\Delta}\sum_{i=1}^{N_\Delta}\sum_{l=1}^L
\big|\Delta^l\big(t^i_\Delta,\mathbf x^i_\Delta,\mathbf u^{\boldsymbol\theta}_{(n)}(t^i_\Delta,\mathbf x^i_\Delta)\big)\big|^2,
\quad (t^i_\Delta,\mathbf x^i_\Delta)\in[t_0,t_{\rm f}]\times\Omega',\ i=1,\dots,N_\Delta,
\\[1.5ex]
\mathcal L_{\rm i}(\boldsymbol\theta) ={}& \frac1{N_{\rm i}}\sum_{i=1}^{N_{\rm i}}\sum_{l_{\rm i}=1}^{L_{\rm i}}
\big|\mathsf I^{l_{\rm i}}\big(\mathbf x^i_{\rm i},\mathbf u_{(n_{\rm i})}^{\boldsymbol\theta}(t_0,\mathbf x^i_{\rm i})\big)\big|^2,
\quad \mathbf x^i_{\rm i}\in\Omega',\ i=1,\dots,N_{\rm i},
\\[1.5ex]
\mathcal L_{\rm sb}(\boldsymbol\theta) ={}& \frac1{N_{\rm sb}}\sum_{i=1}^{N_{\rm sb}}\sum_{l_{\rm sb}=1}^{L_{\rm sb}}
\big|\mathsf B_{\rm s}^{l_{\rm sb}}\big(t^i_{\rm sb},\mathbf x^i_{\rm sb},\mathbf u_{(n_{\rm sb})}^{\boldsymbol\theta}(t^i_{\rm sb},\mathbf x^i_{\rm sb})\big)\big|^2,
\\ & (t^i_{\rm sb},\mathbf x^i_{\rm sb})\in[t_0,t_{\rm f}]\times\p\Omega',\ i=1,\dots,N_{\rm sb}.
\end{align*}
Moreover, the loss function for an initial value problem for a system of ordinary differential equations
has no component~$\mathcal L_{\rm sb}$.

Although physics-informed neural networks give a promising way to approximate solutions of boundary value problem, 
they can fail, e.g., for initial--boundary value problems with long time intervals~\cite{wang23a}. 
One way to overcome the problem for long-term integration is to learn mappings from initial data
to the associated solutions of the corresponding initial--boundary value problems. 
This can be achieved within the DeepONet framework.

\subsection{Deep operator neural networks with soft constraints (sONets)}

DeepONets are designed to approximate (non)linear operators between compact subsets of (finite- or infinite-dimen\-sional) Banach spaces, 
which are usually functional spaces.
The DeepONet architecture that is commonly considered in the literature
includes two main components, the \emph{branch net} and the \emph{trunk net}. 
The branch and trunk nets are respectively responsible for encoding the dependence of the output function 
on the input parameter functions (typically restricted to fixed finite subsets of their domains) 
and its arguments. 
Convolving the outcomes of the branch and trunk networks 
with a trainable constant rank-two tensor gives the output of the DeepONet. 

We specify abstract DeepONets to the particular case of mappings between boundary data 
and solutions of the corresponding boundary value problems for a fixed system of differential equations. 
More specifically, we consider a parameterized family of boundary value problems of the general form~\eqref{eq:GenBVP},
\begin{subequations}\label{eq:GenBVPParamFamily}
\begin{gather}\label{eq:GenSystemOfDEsParamFamily}
\Delta^l(\mathbf x,\mathbf u_{(n)})=0,\quad l=1,\dots, L,\qquad \mathbf x\in\Omega,
\\[1ex]\label{eq:GenBCsParamFamily}
\mathsf B ^{l_{\rm b}}(\mathbf x,\mathbf u_{(n_{\rm b})},\boldsymbol\varphi)=0,\quad l_{\rm b}=1,\dots, L_{\rm b},\qquad \mathbf x\in\p\Omega,
\end{gather}
\end{subequations}
where the boundary conditions are parameterized by a tuple of parameter functions 
$\boldsymbol\varphi=\boldsymbol\varphi(\mathbf x)=(\varphi^1,\dots,\varphi^{L_{\rm p}})$ 
running through a compact (with respect to a metric) set~$\mathcal M$ of sufficiently smooth function $L_{\rm p}$-tuples with domain~$\p\Omega$.%
\footnote{%
In general, the boundary value operator~$\mathsf B$ may involve derivatives of~$\boldsymbol\varphi$ with respect to~$\mathbf{x}$ as well. 
Moreover, the system~\eqref{eq:GenSystemOfDEsParamFamily} may also be parameterized in a similar way by functions 
with domain~$\Omega$.
}

To obtain the tuple~$\boldsymbol\Phi$ of parameters encoding the varying boundary data,
we evaluate the function tuple $\boldsymbol\varphi$ at the set of $N_{\rm s}$ sensor points 
$\mathcal S:=\big\{\mathbf x_{\rm s}^i\big\}_{i=1}^{N_{\rm s}}\subset\p\Omega$,
$\boldsymbol\Phi:=(\boldsymbol\varphi(\mathbf x_{\rm s}^1),\ldots,\boldsymbol\varphi(\mathbf x_{\rm s}^{N_{\rm s}}))$.%
\footnote{\label{fnt:DeepONetParameters}
This is not the only possible choice for the parameter tuple~$\boldsymbol\Phi$ to encode the boundary data. 
Indeed, either the set~$\mathcal M$ is parameterized by a finite number of parameters, or  
we can approximate elements of~$\mathcal M$ by elements of a family~$\mathcal M'$ 
of functions with a finite number of parameters.
These parameters can be used directly to encode the boundary conditions for the network 
instead of values of functions from~$\mathcal M$ at the sensor points. 
This is always the case for initial or boundary value conditions for ordinary differential equations 
due to their parameterization at most by constants. 
}
Then a standard DeepONet $\mathcal G^{\boldsymbol\theta}$ takes the form~\cite{lu21a}
\begin{equation}
    \mathcal G^{\boldsymbol\theta}(\boldsymbol\Phi)(\mathbf x) = 
    \sum_{i,j}{\breve{\boldsymbol\theta}}_{ij}\mathcal B^{\hat{\boldsymbol\theta}}_i(\Phi)\mathcal T^{\check{\boldsymbol\theta}}_j(\mathbf x),
\end{equation}
where $\mathcal B^{\hat{\boldsymbol\theta}}_i$ and $\mathcal T^{\check{\boldsymbol\theta}}_j$ are the embedded outputs of the branch and trunk nets, 
and $\boldsymbol\theta=(\breve{\boldsymbol\theta},\hat{\boldsymbol\theta},\check{\boldsymbol\theta})$ 
is the complete tuples of network weights. 

To define the loss function for training the DeepONet $\mathcal G^{\boldsymbol\theta}$,
in addition to sampling collocation points in~$\Omega$ and~$\p\Omega$, 
one needs to sample function tuples $\boldsymbol\varphi^1$, \dots, $\boldsymbol\varphi^{N_{\rm f}}$ in the set~$\mathcal M$.
Denote $\boldsymbol\Phi^{l_{\rm f}}:=(\boldsymbol\varphi^{l_{\rm f}}(\mathbf x_{\rm s}^1),\ldots,\boldsymbol\varphi^{l_{\rm f}}(\mathbf x_{\rm s}^{N_{\rm s}}))$, 
$l_{\rm f}=1,\dots,N_{\rm f}$. 
The parts of the physics-informed loss \eqref{eq:compositeLossFunction} are modified as follows:
\begin{align*}
\mathcal L_\Delta(\boldsymbol\theta) &= \frac1{N_{\rm f}}\frac1{N_\Delta}\sum_{l_{\rm f}=1}^{N_{\rm f}}\sum_{i=1}^{N_\Delta}\sum_{l=1}^L
\big|\Delta^l\big(\mathbf x^i_\Delta, \mathcal G^{\boldsymbol\theta}(\boldsymbol\Phi^{l_{\rm f}})_{(n)}(\mathbf x^i_\Delta)\big)\big|^2,
\\[1.5ex]
\mathcal L_{\rm b}(\boldsymbol\theta) &= \frac1{N_{\rm f}}\frac1{N_{\rm b}}\sum_{l_{\rm f}=1}^{N_{\rm f}}\sum_{i=1}^{N_{\rm b}}\sum_{l_{\rm b}=1}^{L_{\rm b}}
\big|\mathsf B^{l_{\rm b}}\big(\mathbf x^i_{\rm b},\mathcal G^{\boldsymbol\theta}(\boldsymbol\Phi^{l_{\rm f}})_{(n_{\rm b})}(\mathbf x^i_{\rm b})\big)\big|^2.
\end{align*}

Modifications of the loss function that are required for more specific value problems are obvious. 

The above architecture overcomes some of the problems of the physics-informed neural network. 
However, in this setup DeepONets still need to learn initial and/or boundary conditions of the problems under consideration. In the following, we refer to this soft-constraint deep operator network approach as sONets.

\begin{figure}[t]
    \centering
    \includegraphics{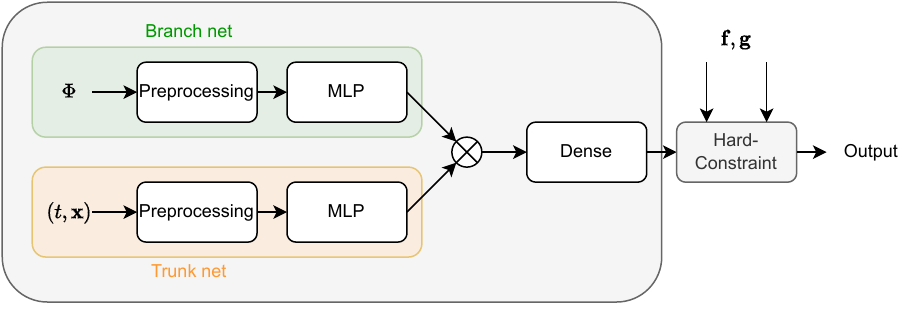}
    \caption{Overview of the used neural network architecture. Here,  multilayer perceptron is abbreviated by MLP.}
    \label{fig:network_architecture}
\end{figure}

\subsection{Deep operator neural networks with hard constraints (hONets)}\label{sec:hONets}

Here, we discuss DeepONets with hard constraints, in the following referred to as hONets, by way of the example of a family of initial--boundary value problems 
of the form~\eqref{eq:GenIBVP} 
for systems of evolution equations with Dirichlet boundary conditions.%
\footnote{%
See examples of possible hard-constraint ansatzes for this and other kinds of boundary value problems 
in \cite{laga98a}, \cite{mose23a} and Sections~\ref{sec:GravityPendulumWithDamping}, \ref{sec:PoissonEq} and~\ref{sec:WaveEq}. 
There is a great variety of such ansatzes even for a specific boundary value problem, 
cf.\ Remark~\ref{rem:HardConstraintForIBVP_OtherPossibilities}.
We plan to discuss selecting them in a forthcoming paper.
}
Thus, 
\begin{gather}\label{eq:IBVPforSystemOfEvolEqsData}
L=q,\quad 
\Delta^l=u^l_t-\tilde\Delta^l(t,\mathbf x,\mathbf u_{(\mathbf x,n)}),\quad 
\mathsf I = \mathbf u|_{t=t_0}^{}-\mathbf f(\mathbf x),\quad 
\mathsf B_{\rm s} = \mathbf u-\mathbf g(t,\mathbf x), 
\end{gather}
and hence the parameter-function tuple is $\boldsymbol\varphi=(\mathbf f,\mathbf g)$ with 
$\mathbf f\colon(\Omega'\cup\p\Omega')\to\mathbb R^q$ and $\mathbf g\colon[t_0,t_{\rm f}]\times\p\Omega'\to\mathbb R^q$.
The consistency of initial and boundary conditions means that 
$\mathbf f(\mathbf x)=\mathbf g(t_0,\mathbf x)$, $\mathbf x\in\p\Omega'$. 
The set of sensor points $\mathcal S$ naturally splits into two subsets, 
the subset $\mathcal S_{\rm i}:=\big\{\mathbf x_{\rm s,i}^i\big\}_{i=1}^{N_{\rm s,i}}\subset\Omega'$ 
of sensor points for the initial conditions and 
the subset $\mathcal S_{\rm sb}:=\big\{\mathbf x_{\rm s,sb}^i\big\}_{i=1}^{N_{\rm s,sb}}\subset[t_0,t_{\rm f}]\times\p\Omega'$ 
of sensor points for the spatial boundary conditions,  
$N_{\rm s}=N_{\rm s,i}+N_{\rm s,sb}$.
Then $\boldsymbol\Phi=(\boldsymbol\Phi_{\rm i},\boldsymbol\Phi_{\rm sb})$, 
where $\boldsymbol\Phi_{\rm i}:=\mathbf f(\mathcal S_{\rm i})$ and $\boldsymbol\Phi_{\rm sb}:=\mathbf g(\mathcal S_{\rm sb})$.

We propose to implement hard constraints 
either directly in the architecture of the DeepONet, see Fig. \ref{fig:network_architecture}, 
or in the loss function, such that initial and boundary conditions are exactly met.

Consider a DeepONet $\mathcal G^{\boldsymbol\theta}(\mathbf f,\mathbf g;\boldsymbol\Phi)(t,\mathbf x)$ with a hard-constraint layer 
that enforces the initial and spatial boundary conditions. 
Thus, the final output is, e.g.,
\begin{gather}\label{eq:HardConstraintForIBVP}
\mathcal G^{\boldsymbol\theta}(\mathbf f,\mathbf g;\boldsymbol\Phi)(t,\mathbf x)=
F_{\rm i}(t,\mathbf x)\mathbf f(\mathbf x)
+F_{\rm sb}(t,\mathbf x)
+F_{\rm nn}(t,\mathbf x)\mathcal G_{\mathrm t}^{\boldsymbol\theta}(\mathbf f,\mathbf g;\boldsymbol\Phi)(t,\mathbf x),
\end{gather}
where $(t,\mathbf x)\in[t_0,t_{\rm f}]\times(\Omega'\cup\p\Omega')$, 
$\mathcal G_{\mathrm t}^{\boldsymbol\theta}$ is the trainable part of the network $\mathcal G^\theta$,
and $F_{\rm i}$, $F_{\rm sb}$ and~$F_{\rm nn}$ are fixed sufficiently smooth functions with domain $[t_0,t_{\rm f}]\times(\Omega'\cup\p\Omega')$ 
that satisfy the conditions 
\begin{subequations}\label{eq:condition_Fs}
\begin{gather}
F_{\rm i}(t_0,\mathbf x)=1,\quad  F_{\rm nn}(t_0,\mathbf x)=0,\quad \frac{\p F_{\rm nn}}{\p t}(t_0,\mathbf x)\ne 0,\quad \mathbf x\in\Omega',
\\
F_{\rm sb}(t,\mathbf x)=\mathbf g(t,\mathbf x)-F_{\rm i}(t,\mathbf x)\mathbf f(\mathbf x),\quad 
F_{\rm nn}(t,\mathbf x)=0,\quad (t,\mathbf x)\in[t_0,t_{\rm f}]\times\p\Omega',
\\
F_{\rm nn}(t,\mathbf x)\ne0,\quad (t,\mathbf x)\in(t_0,t_{\rm f}]\times\Omega'. 
\end{gather}
\end{subequations}
This way we guarantee 
\begin{gather}\label{eq:HardConstraintForIBVPProperties}
\begin{split}
&\mathcal G^{\boldsymbol\theta}(\mathbf f,\mathbf g;\boldsymbol\Phi)(t_0,\mathbf x)=\mathbf f(\mathbf x),\quad \mathbf x\in\Omega',
\\
&\mathcal G^{\boldsymbol\theta}(\mathbf f,\mathbf g;\boldsymbol\Phi)(t,\mathbf x)=\mathbf g(t,\mathbf x),\quad (t,\mathbf x)\in[t_0,t_{\rm f}]\times\p\Omega'.
\end{split}
\end{gather}
Moreover, in view of Hadamard's lemma, see, e.g.,~\cite[Theorem~3.2.15]{kunz15a}, 
any function that satisfies the conditions~\eqref{eq:HardConstraintForIBVPProperties}
can be represented in the form~\eqref{eq:HardConstraintForIBVP}. 
Therefore, the network $\mathcal G^{\boldsymbol\theta}$
can be trained over the neural network parameters~$\boldsymbol\theta$ using the one-component loss function 
including only the equation loss, $\mathcal L(\boldsymbol\theta)=\mathcal L_\Delta(\boldsymbol\theta)$.
This constitutes the main advantage of hard-constrained DeepONets over soft-constrained ones. 

In contrast to the time interval $[t_0,t_{\rm f}]$, 
the geometry of the space domain $\Omega'$ can be complicated, 
and then it may be difficult to take the spatial boundary conditions into account as a part of a hard constraint. 
In this case, one can construct a hard constraint on the basis of the initial conditions only
and train the network using both the system of differential equations and the boundary conditions 
as soft constraints. 
In other words, the ansatz~\eqref{eq:HardConstraintForIBVP} can be replaced by 
\[
\mathcal G^{\boldsymbol\theta}(\mathbf f;\boldsymbol\Phi)(t,\mathbf x)=
F_{\rm i}(t,\mathbf x)\mathbf f(\mathbf x)
+F_{\rm nn}(t,\mathbf x)\mathcal G_{\mathrm t}^{\boldsymbol\theta}(\mathbf f;\boldsymbol\Phi)(t,\mathbf x),
\]
where $(t,\mathbf x)\in[t_0,t_{\rm f}]\times(\Omega'\cup\p\Omega')$, 
$\mathcal G_{\mathrm t}^{\boldsymbol\theta}$ is the trainable part of the network $\mathcal G^\theta$,
and $F_{\rm i}$ and~$F_{\rm nn}$ are fixed sufficiently smooth functions with domain $[t_0,t_{\rm f}]\times(\Omega'\cup\p\Omega')$ 
that satisfy the conditions 
\begin{gather*}
F_{\rm i}(t_0,\mathbf x)=1,\quad  F_{\rm nn}(t_0,\mathbf x)=0,\quad \frac{\p F_{\rm nn}}{\p t}(t_0,\mathbf x)\ne 0,\quad \mathbf x\in\Omega',
\\
F_{\rm nn}(t,\mathbf x)\ne0,\quad (t,\mathbf x)\in(t_0,t_{\rm f}]\times(\Omega'\cup\p\Omega'). 
\end{gather*}
The corresponding loss function is 
$\mathcal L(\boldsymbol\theta) = 
 \lambda_{\Delta}\mathcal L_\Delta(\boldsymbol\theta) 
+\lambda_{\rm i}\mathcal L_{\rm i}(\boldsymbol\theta)$.

\begin{remark}\label{rem:HardConstraintForIBVP_OtherPossibilities}
Other forms of hard constraints that are more specific or more general than \eqref{eq:HardConstraintForIBVP} can be considered. 
In particular, the function $F_{\rm i}$ can be assumed to depend only on~$t$ or even to be the constant function~1 
or, on the other hand, to be a $q\times q$ matrix function with $F_{\rm i}(t_0,\mathbf x)$ equal to the $q\times q$ unit matrix.
\end{remark}

It often happens that the solution of the problem~\eqref{eq:GenIBVP} is not well approximated by a PINN. 
In this case, one can try to partition the entire interval $[t_0,t_{\rm f}]$ 
into subintervals $[t_{j-1},t_j]$, $j=1,\dots,k$ with $t_0<t_1<\dots<t_k=t_{\rm f}$, $k>0$ 
and solve the sequence of initial--boundary value problems on these subinvervals 
using the evaluation of the solution of the $j$th problem at $t=t_j$ as 
the initial condition for the $(j+1)$th problem.  
This is the essence of the time-stepping procedure. 
Of course, the simplest choice here is given by the partition into intervals of the same length $\Delta t$.
The time-stepping can be efficiently realized within the framework of DeepONets 
considering $t_0$ as one more parameter of the network to be used, 
training the network for the network over the interval $[t_0,t_0+\Delta t]$ 
and substituting $t_0+(j-1)\Delta t$ instead of~$t_0$ in the $j$th step of the time-stepping procedure. 
If the system of differential equations in~\eqref{eq:GenIBVP} is invariant with respect to translations in~$t$ 
and the parameter function~$\mathbf g$ in the spatial boundary conditions does not depend on~$t$, 
then the time-stepping procedure is even simpler since it does not require 
modifying the network by extending the parameter set of the network with~$t_0$.

The approximate solution of the problem~\eqref{eq:GenIBVP} in the particular setting~\eqref{eq:IBVPforSystemOfEvolEqsData} 
using the time-stepping procedure for the modified hard-constrained DeepONet~\eqref{eq:HardConstraintForIBVP} 
with the additional parameter~$t_0$ takes the form
\begin{gather*}
\Gamma(t,x)=\mathcal G^\theta(t_{j-1},\mathbf f^{j-1},\mathbf g;\boldsymbol\Phi^j)(t,\mathbf x),\quad 
(t,\mathbf x)\in[t_{j-1},t_j]\times(\Omega'\cup\p\Omega'),\ j=1,\dots,k,
\end{gather*}
where $\mathbf f^0:=\mathbf f$, 
$\mathbf f^j:=\mathcal G^\theta(t_{j-1},\mathbf f^{j-1},\mathbf g;\boldsymbol\Phi^j)(t_j,\cdot)$, $j=1,\dots,k-1$, 
\[
\mathcal S_{\rm i}^j:=\big\{\mathbf x_{\rm s,i}^{j,i}\big\}_{i=1}^{N_{\rm s,i}^j}\subset\Omega', \quad
\mathcal S_{\rm sb}^j:=\big\{\mathbf x_{\rm s,sb}^{j,i}\big\}_{i=1}^{N_{\rm s,sb}^j}\subset[t_{j-1},t_j]\times\p\Omega'
\]
are the sets of sensor points for the initial and the spacial boundary conditions of the $j$th problem, respectively,
$\boldsymbol\Phi^j=(\boldsymbol\Phi_{\rm i}^j,\boldsymbol\Phi_{\rm sb}^j)$ 
with $\boldsymbol\Phi_{\rm i}^j:=\mathbf f(\mathcal S_{\rm i}^j)$ and $\boldsymbol\Phi_{\rm sb}^j:=\mathbf g(\mathcal S_{\rm sb}^j)$, 
$j=1,\dots,k$. 
For simplicity, one can choose $\mathcal S_{\rm i}^1=\dots=\mathcal S_{\rm i}^k:=\mathcal S_{\rm i}$.
When using intervals of the same length $\Delta t$, 
one can obtain the sensor-point sets $\mathcal S_{\rm sb}^j$, $j=2,\dots,k$ 
by means of successively shifting the first sensor-point set $\mathcal S_{\rm sb}^1$ in~$\Delta t$ in the $t$-direction. 
In view of~\eqref{eq:HardConstraintForIBVPProperties} and the construction of the function~$\Gamma$, 
this function is at least a continuous function on $[t_0,t_{\rm f}]\times(\Omega'\cup\p\Omega')$.
Moreover, a greater order of smoothness of the function~$\Gamma$ can be enforced 
by a proper choice of general ansatz structure, of ansatz coefficients like~$F_{\rm i}$, $F_{\rm sb}$ and $F_{\rm nn}$ 
and of the activation functions. 
This is one of the main advantage of the hard-constraint approach over the soft-constraint one, 
which cannot guarantee even the continuity of the solution obtained by time-stepping, 
see Fig.~\ref{fig:lorenz_compare}b.

\begin{example}[Dirichlet boundary condition in 1D] 
Consider a family of initial--boundary value problems with $d=1$, $q=1$, $\Omega'=[0,1]$ 
and Dirichlet boundary condition for a single evolution equation. 
Thus, $\mathsf I=u(t_0,x)-f(x)$, $\mathsf B_{\rm s}=\big(u(t,0)-g^0(t),u(t,1)-g^1(t)\big)$, 
and the consistency of initial and boundary conditions implies that
$g^0(t_0)=f(0)$ and $g^1(t_0)=f(1)$.
We can set $F_{\rm i}\equiv1$, 
$F_{\rm sb}(t,x)=\big(g^0(t)-f(0)\big)(1-x)+\big(g^1(t)-f(1)\big)x$
and $F_{\rm nn}(t,x)=tx(x-1)$. 
\end{example}

We note that depending on the choice of $(F_{\rm i},F_{\rm sb},F_{\rm nn})$ the outcome of the training can be better or worse.

\begin{remark}\label{rem:HardContraintsForODEs}
The case of systems of ordinary differential equations is specific in the framework of hard-constrained DeepONets in several aspects. 
One of them is the same as for soft-constrained DeepONets, see Footnote~\ref{fnt:DeepONetParameters}, 
i.e., there is no selecting sensor points in this case, and the initial or boundary values can be directly used as network inputs. 
For hard constraints specifically, it is essential that most problems considered for ordinary differential equations 
are purely initial, without boundary conditions. 
This necessitates the modification of the ansatz~\eqref{eq:HardConstraintForIBVP} for this setting. 
More specifically, given a family of initial value problems 
$u^l_t-\tilde\Delta^l(t,\mathbf u)=0$, $l=1,\dots,L$, $\mathbf u|_{t=t_0}^{}=\mathbf f$, 
where $\mathbf f$ is a tuple of constant parameters, a possible replacement for~\eqref{eq:HardConstraintForIBVP} is  
\begin{gather}\label{eq:HardConstraintForIVP}
\mathcal G^{\boldsymbol\theta}(\mathbf f)(t)=
F_{\rm i}(t)\mathbf f+F_{\rm nn}(t)\mathcal G_{\mathrm t}^{\boldsymbol\theta}(\mathbf f)(t),
\end{gather}
where $t\in[t_0,t_{\rm f}]$, 
$\mathcal G_{\mathrm t}^{\boldsymbol\theta}$ is the trainable part of the network $\mathcal G^\theta$,
and $F_{\rm i}$ and~$F_{\rm nn}$ are fixed sufficiently smooth functions with domain $[t_0,t_{\rm f}]$ 
that satisfy the conditions 
\begin{gather*}
F_{\rm i}(t_0)=1,\quad  F_{\rm nn}(t_0)=0,\quad \frac{\p F_{\rm nn}}{\p t}(t_0)\ne 0,\quad 
F_{\rm nn}(t)\ne0,\ t\in(t_0,t_{\rm f}]. 
\end{gather*}
Note that here the parameter tuple~$\boldsymbol\Phi$ coincides with~$\mathbf f$ and is thus omitted. 
Pure boundary value problems for systems of ordinary differential equations 
also require specific hard constraints, see Section~\ref{sec:PoissonEq}.
\end{remark}

\section{Results}\label{sec:Results}

In this section, we demonstrate that DeepONets with hard constraints perform better compared to DeepONets with soft constraints.
We first consider three examples of (systems of) ordinary differential equations (ODEs), 
\begin{itemize}\itemsep=0ex
\item
a gravity pendulum with an external damping force in Section~\ref{sec:GravityPendulumWithDamping}, 
\item
the Lorenz--1963 model in Section~\ref{sec:Lorenz1963} and 
\item
a one-dimensional Poisson equation in Section~\ref{sec:PoissonEq},
\end{itemize}
and then two examples of partial differential equations (PDEs),
\begin{itemize}\itemsep=0ex
\item
one-dimensional linear wave equation in Section~\ref{sec:WaveEq} and
\item
the Korteweg--de Vries (KdV) equation in Section~\ref{sec:KdV}. 
\end{itemize}

In Table~\ref{tab:netset} we give the details about the setups of each neural network. 
As the numbers of layers and of units per layers in the corresponding perceptrons, 
we indicate those for each of the trunk and branch subnetworks. 
They are supplemented with the layer applying the linear activation to the tensor product of the trunk and branch outputs 
and, for hONets, additional output layers implementing hard constraints in the network architecture.

\begin{table}[!ht]
\centering
\begin{tabular}{l||c||c||c||c||c}
  Neural network setup & pendulum  & Lorenz--1963 & Poisson & wave & KdV \\ \hline\hline
Sensor points / Parameters   & 2 & 3 & 2 & 100 & 100       \\ \hline 
Collocation points  & $10^4$ & $5\cdot10^4$ & $10^4$ & $10^6$ & $2\cdot10^5$       \\ \hline 
Batch size  & $10^3$ & $10^3$ & $10^3$ & $10^5$ & $2\cdot10^4$       \\ \hline 
Indep. $|$ dep. vars.  & $t\mid x$ & $t\mid x,y,z$ & $x\mid u$ & $t,x\mid u$ & $t,x\mid u$ \\ \hline 
Training domains   & $[0,1]$ & $[0,0.2]$ & $[0,1]$ & $[0,1]$ & $[0,1]$, $[0,10]$       \\ \hline 
MLP (units $\mid$ layers) & $40\mid4$ & $60\mid5$ & $60\mid5$ & $60\mid5$ & $60\mid5$
\end{tabular}
\caption{Neural network setups for the experiments carried out in Section~\ref{sec:Results}.}\label{tab:netset}
\end{table}

For the sake of comparison, all the demonstrated results (except the KdV equation, see Section~\ref{sec:KdV})
were obtained within the same training settings, 
which are given by the Adam optimizer with parameters $\beta_1=0.95$ and $\beta_2=0.99$ and the exponentially decaying learning rate starting at $10^{-2}$ using $20\times\mathsf B$ decay steps with the decay rate of 0.95, where $\mathsf B$ is the number of batches per epoch. These parameters have been determined through hyper-parameter tuning. 
Each training was 5000 epochs. 
The codes for each example are published on GitHub\footnote{\url{https://github.com/RudigerBrecht/hardConstraintDeepONet}} .

To quantify the error, we will use the normalized root mean square error (NRMSE) given by
\begin{gather*}
    \text{NRMSE}(u,u_{\rm ref})=\left.\sqrt{\frac{1}{N}\sum_{i=1}^N(u_i-u_{{\rm ref},i})^2}\right/
    \sqrt{\frac{1}{N}\sum_{i=1}^N(u_{{\rm ref},i})^2}\,.
\end{gather*}
Then we calculate the error reduction of $\text{NRMSE}$ over $\text{NRMSE}_0$ by
$\dfrac{|\text{NRMSE}-\text{NRMSE}_0|}{|\text{NRMSE}_0|}\cdot100\%$.

\begin{remark}
Both the soft- and hard-constraint methods can be further tuned. For soft-constrained DeepONets, the loss weights associated with the initial, boundary and residual values may be adjusted to optimize the training. 
In \cite{wang2022improved}, the authors experimented with changing the weights of the loss components and modifying the architecture of DeepONets. 
They found that for the advection equation, the tuning of weights does not reduce the error significantly, however changing the architecture does. 
For hard-constrained DeepONets, it is reasonable to expect further improvements from the choice of ansatzes for the hard constraints. 
We tried out various, specially selected functions as coefficients in such ansatzes, some of which improved the outcome, 
see Section~\ref{sec:GravityPendulumWithDamping}. 
To have a fair comparison, however, we used the simplest weights for the loss and the simplest function for the hard constraints.
\end{remark}

\begin{remark}
There are two options for implementing a hard constraint within the hONet approach. The first option is to embed it as the last layer in the neural network architecture. The second option is to use the hard constraint as a modifier for the output of the neural network before substituting the result into the loss function.
\end{remark}

\begin{notation*}
Throughout this section, $\mathcal G_{\mathrm t}^{\boldsymbol\theta}$ denotes the trainable part of the network $\mathcal G^\theta$,
and $t_{\rm n}:=(t-t_0)/(t_{\rm f}-t_0)$.
\end{notation*}

\subsection{Gravity pendulum with damping}\label{sec:GravityPendulumWithDamping}

The behavior of a gravity pendulum subjected to an external damping force can be characterized 
by the following initial value problem, cf.\ \cite[Section~2]{wang23a}:
\begin{subequations}\label{eq:GravityPendulumWithDamping}
\begin{gather}\label{eq:GravityPendulumWithDampingA}
\frac{{\rm d}^2x}{{\rm d}t^2} + \frac bm\frac{{\rm d}x}{{\rm d}t} + \frac gL\sin(x)=0,\quad  t\in[t_0,t_{\rm f}], 
\\ \label{eq:GravityPendulumWithDampingB}
x(t_0)=x_0,\quad \frac{{\rm d}x}{{\rm d}t}(t_0)=x_{t,0}.
\end{gather}
\end{subequations}
Here 
$m$ is the bob mass, 
$l$ is the rod length,
$b$ is the damping coefficient and
$g$ is the gravitational acceleration constant. 
We take $m=1$, $L=1$, $b = 0.05$ and $g = 9.81$. 

Two constant parameters encode varying initial data in the DeepONet, $x_0$ and~$x_{t,0}$. 
This is why as for any initial or boundary value problem for a system of ordinary differential equations, 
there is no selecting sensor points here, and $\boldsymbol\Phi=(x_0,x_{t,0})$, 
see Remark~\ref{rem:HardContraintsForODEs}.
Following Section~\ref{sec:hONets}, we consider 
the general form of ansatzes for hard constraint layers that satisfies the properties~\eqref{eq:condition_Fs} 
in this example is given by
\begin{gather}\label{HardConstraintWith1stOrderDersCoeffsGenForm}
x(t)\approx\mathcal G^{\boldsymbol\theta}(x_0,x_{t,0})(t)=
F_{\rm i0}(t)x_0 +F_{\rm i1}(t)x_{t,0}+F_{\rm nn}(t)\mathcal G_{\mathrm t}^{\boldsymbol\theta}(x_0,x_{t,0})(t),
\end{gather}
where $t\in[t_0,t_{\rm f}]$, 
and $F_{\rm i0}$, $F_{\rm i1}$ and~$F_{\rm nn}$ are fixed sufficiently smooth functions with domain $[t_0,t_{\rm f}]$ 
that satisfy at least the conditions 
\begin{gather}\label{eq:ODEHardConstraintWith1stOrderDersCoeffs}
\begin{split}&
F_{\rm i0}(t_0)=1,\quad  F_{\rm i1}(t_0)=0,\quad  F_{\rm nn}(t_0)=0,\\& 
\frac{{\rm d} F_{\rm i0}}{{\rm d}t}(t_0)=0,\quad 
\frac{{\rm d}F_{\rm i1}}{{\rm d}t}(t_0)=1,\quad 
\frac{{\rm d}F_{\rm nn}}{{\rm d}t}(t_0)=0,\\&
\frac{{\rm d}^2 F_{\rm nn}}{{\rm d}t^2}(t_0)\ne0,\quad 
F_{\rm nn}(t)\ne0,\ t\in(t_0,t_{\rm f}]. 
\end{split}
\end{gather}
Each of ansatzes of the form~\eqref{HardConstraintWith1stOrderDersCoeffsGenForm} guarantees that 
any approximate solution $\mathcal G^{\boldsymbol\theta}(x_0,x_{t,0})$ 
obtained by the time-stepping with involvement of this ansatz 
is at least once continuously differentiable on the entire multi-step integration interval
and hence is proper in the weak sense. 
An obvious and simplistic ansatz of the form~\eqref{HardConstraintWith1stOrderDersCoeffsGenForm} is 
that with 
\begin{gather}\label{HardConstraintWith1stOrderDersCoeffs1}
F_{\rm i0}(t)=1,\quad 
F_{\rm i1}(t)=t-t_0,\quad  
F_{\rm nn}(t)=t_{\rm n}^2,
\end{gather}
i.e., 
$x(t)\approx\mathcal G^{\boldsymbol\theta}(x_0,x_{t,0})(t)
=x_0 +(t-t_0)x_{t,0} + t_{\rm n}^2 \mathcal G^{\boldsymbol\theta}_{\rm t}(x_0,x_{t,0})(t)$.
More sophisticated forms of the hard constraint could be investigated, which may improve upon the results reported even further.
In particular, we tested the following choices for the ansatz coefficients:
\begin{gather}\label{HardConstraintWith1stOrderDersCoeffs2}
F_{\rm i0}(t)=1-3t_{\rm n}^{\,2}+2t_{\rm n}^{\,3},\quad
F_{\rm i1}(t)=(t-t_0)(1-t_{\rm n})^2,\quad
F_{\rm nn}(t)=t_{\rm n}^{\,2}(3-2t_{\rm n}),
\\[1ex]\label{HardConstraintWith1stOrderDersCoeffs3}
\begin{split}&   
F_{\rm i0}(t)=1-10t_{\rm n}^{\,3}+15t_{\rm n}^{\,4}-6t_{\rm n}^{\,5},\quad
F_{\rm i1}(t)=(t-t_0)(1-6t_{\rm n}^{\,2}+8t_{\rm n}^{\,3}-3t_{\rm n}^{\,4}),\\&
F_{\rm nn}(t)=10t_{\rm n}^{\,2}-20t_{\rm n}^{\,3}+15t_{\rm n}^{\,4}-4t_{\rm n}^{\,5}.
\end{split}
\end{gather}

Trying to maximally follow the spirit of the hard-constraint approach, 
we can also take into account that setting the values of the function~$x$ and its first derivative~${\rm d}x/{\rm d}t$ 
at a time~$t$ also defines the value of its second derivative~${\rm d}^2x/{\rm d}t^2$ at this time
in view of the equation~\eqref{eq:GravityPendulumWithDamping}. 
The corresponding ansatz for hard constraints is 
\begin{gather}\label{HardConstraintWith2ndOrderDersCoeffsGenForm}
x(t)\approx\mathcal G^{\boldsymbol\theta}(x_0,x_{t,0})(t)=
F_{\rm i0}(t)x_0+F_{\rm i1}(t)x_{t,0}+F_{\rm i2}(t)x_{tt,0}+F_{\rm nn}(t)\mathcal G_{\mathrm t}^{\boldsymbol\theta}(x_0,x_{t,0})(t),
\end{gather}
where $x_{tt,0}:=-bm^{-1}x_{t,0} + gL^{-1}\sin(x_0)$,
$t\in[t_0,t_{\rm f}]$, 
and $F_{\rm i0}$, $F_{\rm i1}$, $F_{\rm i2}$ and~$F_{\rm nn}$ are fixed sufficiently smooth functions with domain $[t_0,t_{\rm f}]$ 
that satisfy at least the conditions 
\begin{gather}\label{eq:ODEHardConstraintWith2ndOrderDersCoeffs}
\begin{split}&
F_{\rm i0}(t_0)=1,\quad  F_{\rm i1}(t_0)=0,\quad  F_{\rm i2}(t_0)=0,\quad  F_{\rm nn}(t_0)=0,
\\& 
\frac{{\rm d}F_{\rm i0}}{{\rm d}t}(t_0)=0,\quad 
\frac{{\rm d}F_{\rm i1}}{{\rm d}t}(t_0)=1,\quad 
\frac{{\rm d}F_{\rm i2}}{{\rm d}t}(t_0)=0,\quad 
\frac{{\rm d}F_{\rm nn}}{{\rm d}t}(t_0)=0,
\\&
\frac{{\rm d}^2F_{\rm i0}}{{\rm d}t^2}(t_0)=0,\quad 
\frac{{\rm d}^2F_{\rm i1}}{{\rm d}t^2}(t_0)=0,\quad 
\frac{{\rm d}^2F_{\rm i2}}{{\rm d}t^2}(t_0)=1,\quad 
\frac{{\rm d}^2F_{\rm nn}}{{\rm d}t^2}(t_0)=0,
\\&
\frac{{\rm d}^3 F_{\rm nn}}{{\rm d}t^3}(t_0)\ne0,\quad 
F_{\rm nn}(t)\ne0,\ t\in(t_0,t_{\rm f}]. 
\end{split}
\end{gather}
As a reasonable choice, we consider the coefficients
\begin{gather}\label{ODEHardConstraintWith2ndOrderDersCoeffs1}
\begin{split}&   
F_{\rm i0}(t)=1-10t_{\rm n}^{\,3}+15t_{\rm n}^{\,4}-6t_{\rm n}^{\,5},\quad
F_{\rm i1}(t)=(t-t_0)(1-6t_{\rm n}^{\,2}+8t_{\rm n}^{\,3}-3t_{\rm n}^{\,4}),\\&
F_{\rm i2}(t)=\tfrac12(t-t_0)^2(1-t_{\rm n})^3,\quad
F_{\rm nn}(t)=10t_{\rm n}^{\,3}-15t_{\rm n}^{\,4}+6t_{\rm n}^{\,5}.
\end{split}
\end{gather}
In addition to the conditions~\eqref{eq:ODEHardConstraintWith2ndOrderDersCoeffs}, 
the coefficients~\eqref{ODEHardConstraintWith2ndOrderDersCoeffs1} and 
their first and second derivatives are equal to zero at the time~$t_{\rm f}$, 
except $F_{\rm nn}(t_{\rm f})=1$. 
This is why if the coefficients~\eqref{ODEHardConstraintWith2ndOrderDersCoeffs1} 
are chosen in~\eqref{HardConstraintWith2ndOrderDersCoeffsGenForm}, then 
the values of the involved functions~$\mathcal G^{\boldsymbol\theta}(x_0,x_{t,0})$ and~$\mathcal G_{\mathrm t}^{\boldsymbol\theta}(x_0,x_{t,0})$
(resp., of their first derivatives or, resp., of their second derivatives) coincide at the time~$t_{\rm f}$.

The results of test computations that are presented in Table~\ref{tab:pendulum} 
clearly show the superiority of the hONet with hard constraint~\eqref{eq:ODEHardConstraintWith2ndOrderDersCoeffs},    
\eqref{ODEHardConstraintWith2ndOrderDersCoeffs1} over the other DeepONets described above in this section, 
especially in the course of using for time-stepping.
Nevertheless, although the ansatz~\eqref{eq:ODEHardConstraintWith2ndOrderDersCoeffs} 
involve the second-order derivative of~$x$, 
it does not guarantee that a solution obtained by time-stepping based on the corresponding hONet 
is twice continuously differentiable globally. 
The reason for this is that for each step interval, the constructed solution satisfies the equation 
only approximately, including the end of the interval, 
and hence its second derivative in general jumps when transitioning to the next step interval. 
We can modify the ansatz~\eqref{eq:ODEHardConstraintWith2ndOrderDersCoeffs} 
to enforce satisfying the equation by the solution at the end point of the training domain, 
\begin{gather}\label{HardConstraintWith2ndOrderDersGenFormMod}
\mathcal G^{\boldsymbol\theta}_{\rm m}(x_0,x_{t,0})(t)=
\mathcal G^{\boldsymbol\theta}(x_0,x_{t,0})(t)-F_{\rm m}(t)\Delta\big(\mathcal G^{\boldsymbol\theta}(x_0,x_{t,0})\big)(t_{\rm f}),
\end{gather}
where 
$\Delta\big(\mathcal G^{\boldsymbol\theta}(x_0,x_{t,0})\big)$ 
is the evaluation of the left-hand side of the equation~\eqref{eq:GravityPendulumWithDamping} at 
the function~$\mathcal G^{\boldsymbol\theta}(x_0,x_{t,0})$, 
and the coefficient~$F_{\rm m}$ satisfies the conditions
\[
F_{\rm m}(t_0)=\frac{{\rm d}F_{\rm m}}{{\rm d}t}(t_0)=\frac{{\rm d}^2F_{\rm m}}{{\rm d}t^2}(t_0)
=F_{\rm m}(t_{\rm f})=\frac{{\rm d}F_{\rm m}}{{\rm d}t}(t_{\rm f})=0,\quad
\frac{{\rm d}^2F_{\rm m}}{{\rm d}t^2}(t_{\rm f})=1.
\]
We can choose $F_{\rm m}(t)=\frac12t_{\rm n}^{\,3}(t-t_{\rm f})^2$.

We can use the basic ansatz~\eqref{HardConstraintWith1stOrderDersCoeffs1} 
and easily construct adaptive hard constraints, e.g., 
by linearly combining the coefficients tuples
\eqref{HardConstraintWith1stOrderDersCoeffs1}--\eqref{HardConstraintWith1stOrderDersCoeffs3} componentwise 
and allowing for training the coefficients of the combination. 
Since the tuples~\eqref{HardConstraintWith1stOrderDersCoeffs2} and~\eqref{HardConstraintWith1stOrderDersCoeffs3} 
showed better results than~\eqref{HardConstraintWith1stOrderDersCoeffs1}, 
for initial experiments we consider convex linear combinations of the components of the former tuples, 
\begin{gather}\label{HardConstraintWith1stOrderDersCoeffsLinComb23}
\begin{split}&   
F_{\rm i0}(t)=a_1(1-10t_{\rm n}^{\,3}+15t_{\rm n}^{\,4}-6t_{\rm n}^{\,5})+(1-a_1)(2t_{\rm n}^{\,3}-3t_{\rm n}^{\,2}+1),\\&
F_{\rm i1}(t)=a_2(t-t_0)(1-6t_{\rm n}^{\,2}+8t_{\rm n}^{\,3}-3t_{\rm n}^{\,4})+(1-a_2)(t-t_0)(t_{\rm n}-1)^2,\\&
F_{\rm nn}(t)=a_3(10t_{\rm n}^{\,2}-20t_{\rm n}^{\,3}+15t_{\rm n}^{\,4}-4t_{\rm n}^{\,5})+(1-a_3)t_{\rm n}^{\,2}(3-2t_{\rm n}),
\end{split}
\end{gather}
where the parameters~$a_1$, $a_2$ and $a_3$ are trainable. 
For comparison, we then take convex linear combinations of the components of all the coefficients tuples
\eqref{HardConstraintWith1stOrderDersCoeffs1}--\eqref{HardConstraintWith1stOrderDersCoeffs3},
\begin{gather}\label{HardConstraintWith1stOrderDersCoeffsLinComb123}
\begin{split}&   
F_{\rm i0}(t)=a_1(1-10t_{\rm n}^{\,3}+15t_{\rm n}^{\,4}-6t_{\rm n}^{\,5})+a_4(2t_{\rm n}^{\,3}-3t_{\rm n}^{\,2}+1)+(1-a_1-a_4),\\&
F_{\rm i1}(t)=a_2(t-t_0)(1-6t_{\rm n}^{\,2}+8t_{\rm n}^{\,3}-3t_{\rm n}^{\,4})+a_5(t-t_0)(t_{\rm n}-1)^2+(1-a_2-a_5)(t-t_0),\\&
F_{\rm nn}(t)=a_3(10t_{\rm n}^{\,2}-20t_{\rm n}^{\,3}+15t_{\rm n}^{\,4}-4t_{\rm n}^{\,5})+a_6t_{\rm n}^{\,2}(3-2t_{\rm n})+(1-a_3-a_6)t_{\rm n}^2,
\end{split}
\end{gather}
where the parameters~$a_1$, \dots $a_6$ are trainable. 

\begin{figure}[!ht]
    \centering
    \includegraphics[width=.9\textwidth]{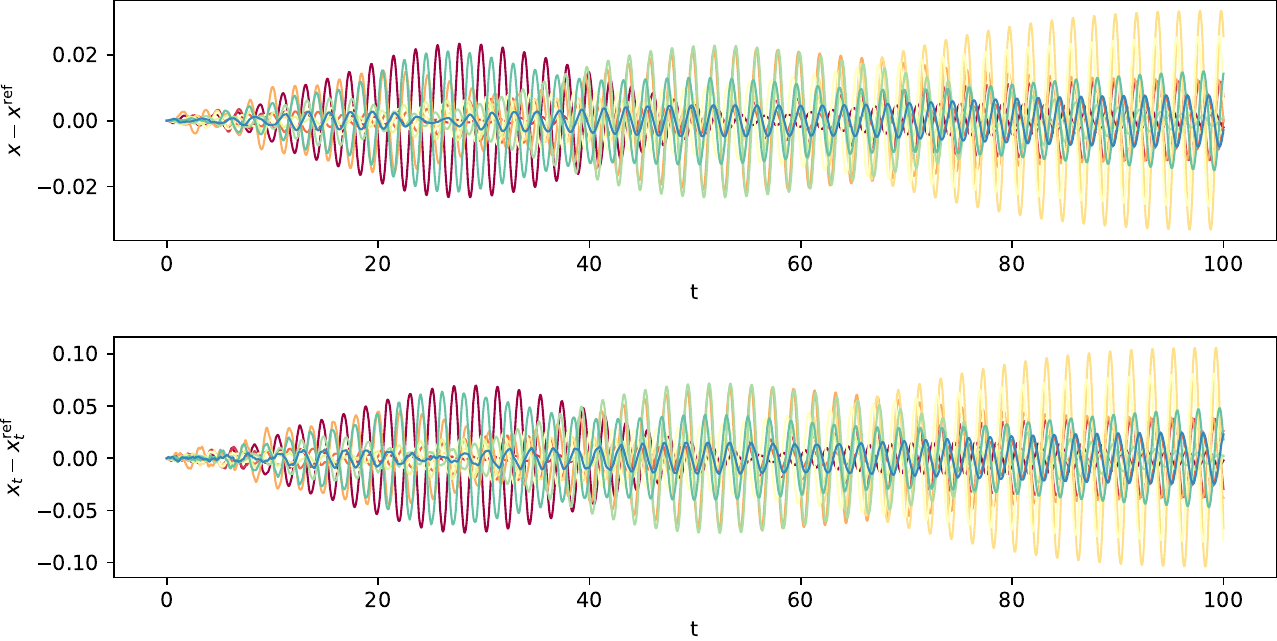}
    \caption{The evolution of pointwise error of 10 ahONet1 training runs in the course of time stepping $100$ times corresponding to the time interval $[0,100]$.}
    \label{fig:pendulum_ahONet1}
\end{figure}

\begin{figure}[!ht]
    \centering
    \includegraphics[width=\textwidth]{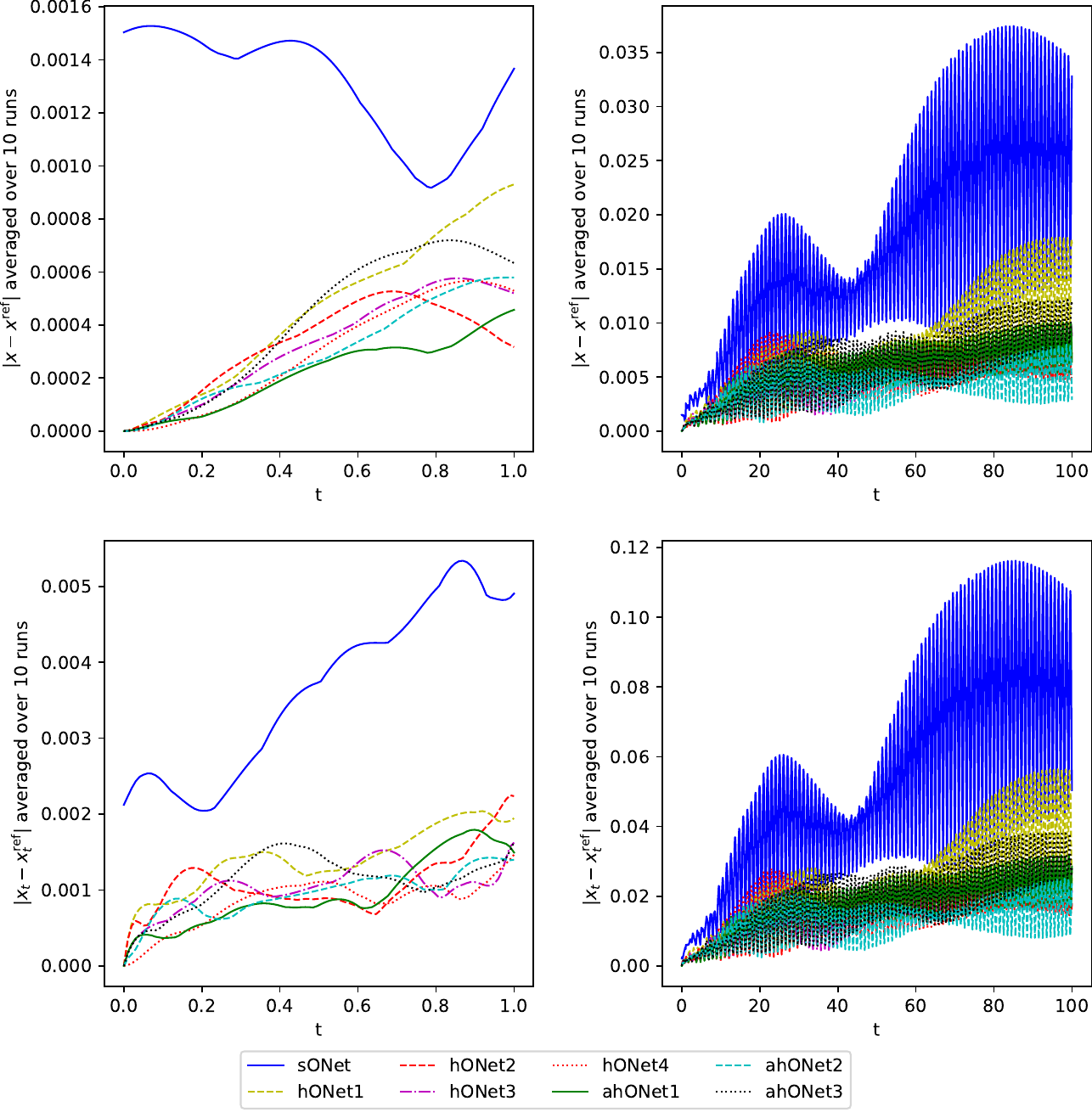}
    \caption{The evolution of absolute error averaged over 10 runs for the damped gravity pendulum. 
    \textit{Left:} Errors over a single time-step, i.e., the training interval $[0,1]$ for the DeepOnets. 
    \textit{Right:} Errors over a total of one hundred time steps, using the time-stepping capabilities of DeepOnets.}
    \label{fig:pendulum_compare}
\end{figure}

In total, we train 
\begin{itemize}\itemsep=0ex
\item
one DeepONet with the soft constraint (referred to as sONet),
\item
four DeepONets with the hard constraints  
that are defined by the general ansatz~\eqref{HardConstraintWith1stOrderDersCoeffsGenForm} 
with the coefficients~\eqref{HardConstraintWith1stOrderDersCoeffs1},  
\eqref{HardConstraintWith1stOrderDersCoeffs2} and~\eqref{HardConstraintWith1stOrderDersCoeffs3} 
and by the general ansatz~\eqref{HardConstraintWith2ndOrderDersCoeffsGenForm} 
with the coefficients~\eqref{ODEHardConstraintWith2ndOrderDersCoeffs1} 
and are referred to as hONet1, hONet2, hONet3 and hONet4, respectively, 
\item
two DeepONets with the same adaptive hard constraint defined 
by the general ansatz~\eqref{HardConstraintWith1stOrderDersCoeffsGenForm} 
with the coefficients~\eqref{HardConstraintWith1stOrderDersCoeffsLinComb23} 
and with the initial values $(0.5,0.5,0.5)$ and $(0.75,0.75,0.75)$ 
of the trainable parameter tuple $(a_1,a_2,a_3)$, which we refer to as ahONet1 and ahONet2,~and
\item
one DeepONet with the adaptive hard constraint defined 
by the general ansatz~\eqref{HardConstraintWith1stOrderDersCoeffsGenForm} 
with the coefficients~\eqref{HardConstraintWith1stOrderDersCoeffsLinComb123} 
and with the same initial value $0.5$ of the trainable parameters~$a_1$,~\dots $a_6$, 
which is referred to as ahONet3.
\end{itemize}
For this purpose, 
we simultaneously sample $10\,000$ values for initial positions~$x_0$, initial velocities~$x_{t,0}$ 
and the time variable~$t$ uniformly within the intervals $[-3, 3]$, $[-3, 3]$ and $[0, 1]$, respectively. 
Then, we evaluate the soft and hard constrained DeepONets for $x_0=x_{t,0}=1$ and compare them after $1$ and $100$ time steps of length~1. 
For each setting, we repeat the computation 10 times and average the computed errors 
since error values essentially vary between training runs, 
see, e.g., Fig.~\ref{fig:pendulum_ahONet1} for ahONet1 results.
In Fig.~\ref{fig:pendulum_compare} we show the differences of absolute errors in the simulation using the soft and hard constraints. 
The normalized root mean square errors are compared in Table~\ref{tab:pendulum}.

\begin{table}[!ht]
\centering
\begin{tabular}{l||c|c||c|c}
Pendulum& \multicolumn{2}{c||}{NRMSE (1 step)}  & \multicolumn{2}{c}{NRMSE (100 steps)}  \\ \hline
       & $x$               &${\rm d}x/{\rm d}t$& $x$               &${\rm d}x/{\rm d}t$\\ \hline
\hline
sONet & $2.0\cdot10^{-3}$ & $1.9\cdot10^{-3}$ & $6.5\cdot10^{-2}$ & $6.7\cdot10^{-2}$ \\ \hline
\hline
hONet1, \eqref{HardConstraintWith1stOrderDersCoeffsGenForm}\&\eqref{HardConstraintWith1stOrderDersCoeffs1}
& $7.6\cdot10^{-4}$ & $7.6\cdot10^{-4}$ & $3.1\cdot10^{-2}$ & $3.2\cdot10^{-2}$ \\ \hline
hONet2, \eqref{HardConstraintWith1stOrderDersCoeffsGenForm}\&\eqref{HardConstraintWith1stOrderDersCoeffs2} 
& $5.3\cdot10^{-4}$ & $5.9\cdot10^{-4}$ & $1.7\cdot10^{-2}$ & $1.8\cdot10^{-2}$ \\ \hline
hONet3, \eqref{HardConstraintWith1stOrderDersCoeffsGenForm}\&\eqref{HardConstraintWith1stOrderDersCoeffs3} 
& $5.4\cdot10^{-4}$ & $5.4\cdot10^{-4}$ & $2.0\cdot10^{-2}$ & $2.1\cdot10^{-2}$ \\ \hline
hONet4, \eqref{HardConstraintWith2ndOrderDersCoeffsGenForm}\&\eqref{ODEHardConstraintWith2ndOrderDersCoeffs1}
&$5.1\cdot10^{-4}$ & $\boldsymbol{4.7\cdot10^{-4}}$ & $1.9\cdot10^{-2}$ & $2.0\cdot10^{-2}$ \\ \hline
ahONet1 & $\boldsymbol{3.8\cdot10^{-4}}$ & $5.3\cdot10^{-4}$ & $2.2\cdot10^{-2}$ & $2.3\cdot10^{-2}$ \\ \hline
ahONet2 & $5.0\cdot10^{-4}$ & $5.2\cdot10^{-4}$ & $\boldsymbol{1.6\cdot10^{-2}}$ & $\boldsymbol{1.7\cdot10^{-2}}$ \\ \hline
ahONet3 & $6.9\cdot10^{-4}$ & $5.9\cdot10^{-4}$ & $2.4\cdot10^{-2}$ & $2.5\cdot10^{-2}$ \\ \hline 
\hline
Error reduction 1 &     61\% &     59\% &     52\% &      52\% \\ \hline
Error reduction 2 &     73\% &     68\% &     73\% &      73\% \\ \hline
Error reduction 3 &     72\% &     71\% &     69\% &      69\% \\ \hline
Error reduction 4 &     74\% & {\bf74\%}&     70\% &      70\% \\ \hline
Error reduction 5 & {\bf81\%}&     71\% &     65\% &      65\% \\ \hline
Error reduction 6 &     75\% &     72\% & {\bf75\%}& {\bf75\%} \\ \hline
Error reduction 7 &     65\% &     68\% &     63\% &      63\% \\ 
\end{tabular}
\vspace{1ex}
\caption{Normalized root mean square error (NRMSE) for the soft and hard constraint simulations 
of the gravity pendulum with damping~\eqref{eq:GravityPendulumWithDamping}, 
and reduction in error obtained by using the hard constraints over the soft constraint, 
which are averaged over 10 runs
after $1$ and $100$ time steps corresponding to the time intervals $[0,1]$ and $[0,100]$.} \label{tab:pendulum}
\end{table}

\begin{remark}\label{rem:SingleEqForGravityPendulum}
In the classical numerical analysis, the initial value problem~\eqref{eq:GravityPendulumWithDamping}
for the single second-order ordinary differential equation~\eqref{eq:GravityPendulumWithDampingA}
is often replaced by an equivalent initial value problem 
for a system of two first-order ordinary differential equations 
using the standard trick of denoting $x_1:=x$ and $x_2:={\rm d}x/{\rm d}t$,
\begin{gather*}
\Delta^1\left(t,\mathbf x,\frac{{\rm d}\mathbf x}{{\rm d}t}\right):=\frac{{\rm d}x_1}{{\rm d}t}-x_2=0,\ \ \!
\Delta^2\left(t,\mathbf x,\frac{{\rm d}\mathbf x}{{\rm d}t}\right):=\frac{{\rm d}x_2}{{\rm d}t}=-\frac bmx_2 -\frac gL\sin(x_1),\ \ \! t\in[t_0,t_{\rm f}],
\\
x_1(t_0)=x_{10},\quad x_2(t_0)=x_{20}.
\end{gather*}
There are two reasons why this replacement is not advantageous when solving this system with neural networks. 
The first reason is that due to using automatic differentiation 
and the absence of discretization error, 
the calculation of higher-order derivatives stops being an issue. 
The second reason is that the transition from a single equation to a system 
makes the equation loss itself composite, 
$\mathcal L_\Delta(\boldsymbol\theta)=\mathcal L_{\Delta^1}(\boldsymbol\theta)+\mathcal L_{\Delta^2}(\boldsymbol\theta)$, 
which needlessly complicates the training. 
Solving the initial value problem~\eqref{eq:GravityPendulumWithDamping}, 
one implicitly enforces exactly the equation ${\rm d}x_1/{\rm d}t=x_2$, 
treating it as a hard constraint 
even when training a network only with soft (from the formal point of view) constraints. 
\end{remark}

\begin{remark}
The damped gravity pendulum is specific in that for long enough simulation times, the solution asymptotically decays to zero. As such, also the error of any meaningful numerical approximation to this solution will get smaller over time, which is not the case for more complicated time-dependent systems of differential equations, where the approximation error of numerical solutions would normally increase over time. We evaluate the obtained solution over time interval $100$ to better understand its dynamics. 
\end{remark}




\subsection{The Lorenz--1963 model}\label{sec:Lorenz1963}

We consider the family of initial value problems of the Lorenz--1963 model~\cite{lore63Ay}  given by 
\begin{subequations}\label{eq:Lorenz1963}
\begin{gather}\label{eq:Lorenz1963system}
\frac{\mathrm{d}x}{\mathrm{d}t}= \sigma (y-x),\quad 
\frac{\mathrm{d}y}{\mathrm{d}t}=x(\rho-z)-y,\quad  
\frac{\mathrm{d}z}{\mathrm{d}t}=xy-\beta z,\quad  t\in[t_0,t_{\rm f}],
\\\label{eq:Lorenz1963IC}
x(t_0)=x_0,\quad y(t_0)=y_0,\quad z(t_0)=z_0, 
\end{gather}
\end{subequations}
where we choose the standard non-dimensional parameters $\sigma=10$, $\beta=\frac{8}{3}$ and $\rho=28$. 
This formulation yields chaotic dynamics. 
The initial conditions~\eqref{eq:Lorenz1963IC} are only parameterized by three constants~$x_0$, $y_0$ and $z_0$, and hence 
$\boldsymbol\Phi=\mathbf f=(x_0,y_0,z_0)$. 
In other words, there are again no selecting sensor points here, and we directly sample values of $(x_0,y_0,z_0)$ 
in a compact subset of~$\mathbb R^3$, see Remark~\ref{rem:HardContraintsForODEs}.

The hard constraint is introduced in the network architecture by
\[
\begin{pmatrix}x(t)\\y(t)\\z(t)\end{pmatrix}
\approx\mathcal G^{\boldsymbol\theta}(x_0,y_0,z_0)(t)
=(1-t_{\rm n})\begin{pmatrix}x_0\\y_0\\z_0\end{pmatrix}
+t_{\rm n}\mathcal G^{\boldsymbol\theta}_{\rm t}(x_0,y_0,z_0)(t).
\]
In contrast to the other examples, 
here the outputs 
of the networks $\mathcal G^{\boldsymbol\theta}_{\rm t}(x_0,y_0,z_0)$ and thus $\mathcal G^{\boldsymbol\theta}(x_0,y_0,z_0)$ 
are not scalars but three-dimensional vectors. 

Analogously to the consideration of gravity pendulum in Section~\ref{sec:GravityPendulumWithDamping}, 
we can take into account that the initial values of the first-order derivatives $(x_{t,0},y_{t,0},z_{t,0})^{\mathsf T}$
are completely defined once the initial values of the unknown functions $(x_0,y_0,z_0)$ are provided, 
\[
x_{t,0}=\sigma (y_0-x_0),\quad
y_{t,0}=x_0(\rho-z_0)-y_0,\quad
z_{t,0}=x_0y_0-\beta z_0.
\]
This leads to the hard-constraint ansatz
\begin{gather*}
\begin{pmatrix}x(t)\\y(t)\\z(t)\end{pmatrix}
\approx\mathcal G^{\boldsymbol\theta}(x_0,y_0,z_0)(t)=
F_{\rm i0}(t)\begin{pmatrix}x_0\\y_0\\z_0\end{pmatrix}
+F_{\rm i1}(t)\begin{pmatrix}x_{t,0}\\y_{t,0}\\z_{t,0}\end{pmatrix}
+F_{\rm nn}(t)\mathcal G_{\mathrm t}^{\boldsymbol\theta}(x_0,x_{t,0})(t),
\end{gather*}
where $t\in[t_0,t_{\rm f}]$, 
and $F_{\rm i0}$, $F_{\rm i1}$ and~$F_{\rm nn}$ are fixed sufficiently smooth functions with domain $[t_0,t_{\rm f}]$ 
that satisfy at least the conditions~\eqref{eq:ODEHardConstraintWith1stOrderDersCoeffs}, 
and thus~%
\eqref{HardConstraintWith1stOrderDersCoeffs1}, 
\eqref{HardConstraintWith1stOrderDersCoeffs2} or~%
\eqref{HardConstraintWith1stOrderDersCoeffs3}
are possible choices for these coefficients. 

We train DeepONets with the soft constraint as well as with hard constraints for the interval $[t_0,t_{\rm f}]$ with $t_0=0$ and $t_{\rm f}=0.2$. 
For the training, we sample $50\,000$ values for $(t,x_0,y_0,z_0)$, 
where each of $t$, $x_0$, $y_0$ and~$z_0$ is sampled uniformly in the intervals $[0, 0.2], [-20,20], [-25,25]$ and $[0, 50]$, respectively.

We again take advantage of the fact that the model under consideration is translation-invariant with respect to~$t$
to use the trained model to perform time-stepping with the step $\Delta t=0.2$, the accuracy of which we use to compare 
the obtained hard-constrained DeepONet with the standard soft-constrained DeepONet for the problem~\eqref{eq:Lorenz1963}. 
We would like to emphasize that due to the above invariance, we do not need to modify the network~$\mathcal G^{\boldsymbol\theta}$ 
by means of introducing $t_0$ as an additional network parameter. 

\begin{figure}[t]
\centering
\begin{subfigure}[b]{0.48\textwidth}
\centering
    \includegraphics[width=\textwidth]{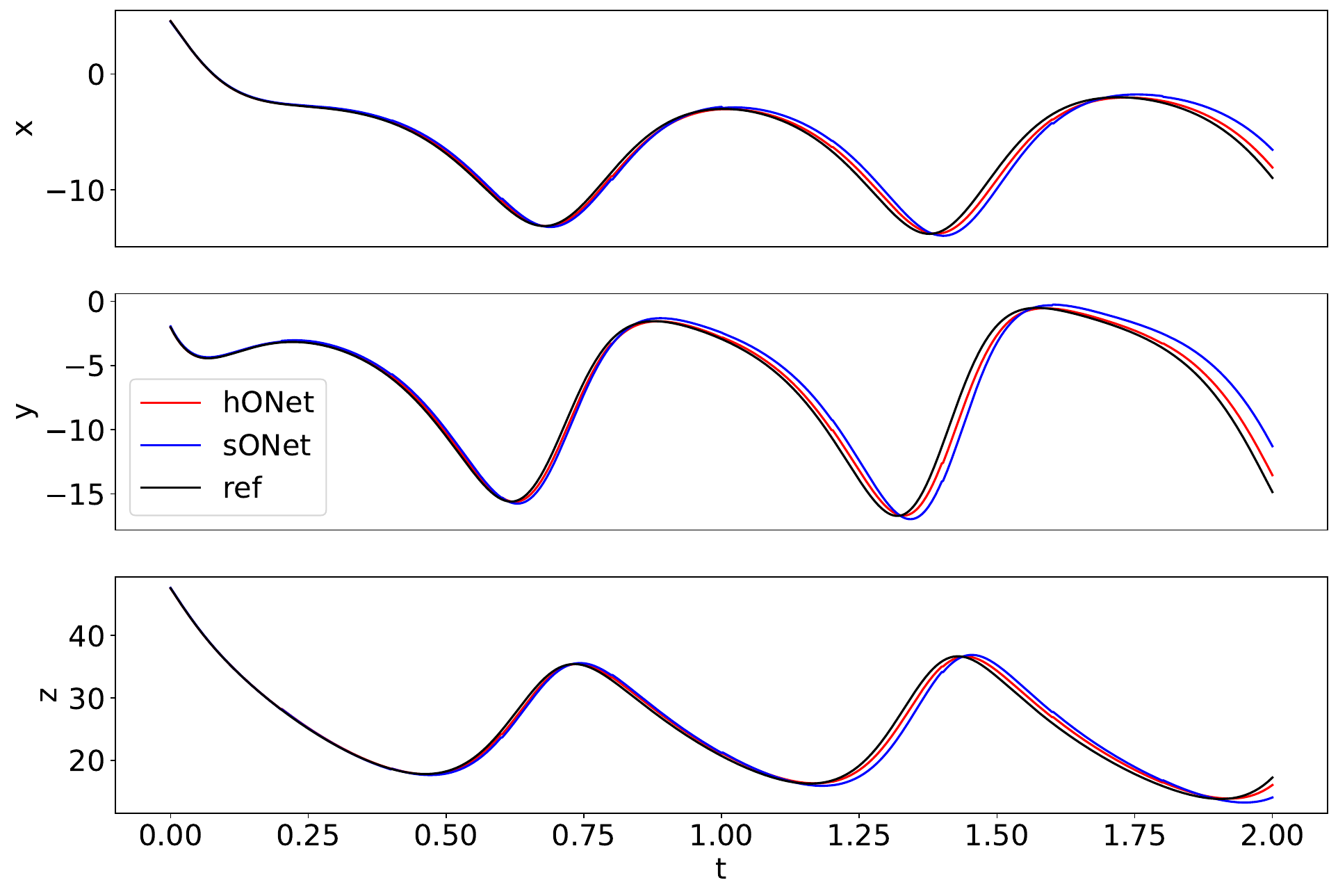}
    \caption{Simulation of a trajectory of the Lorenz--1963 system for $t=2$.  }
    \label{fig:lorenz_compare_good}
\end{subfigure}
\quad
\begin{subfigure}[b]{0.48\textwidth}
\centering
    \includegraphics[width=\textwidth]{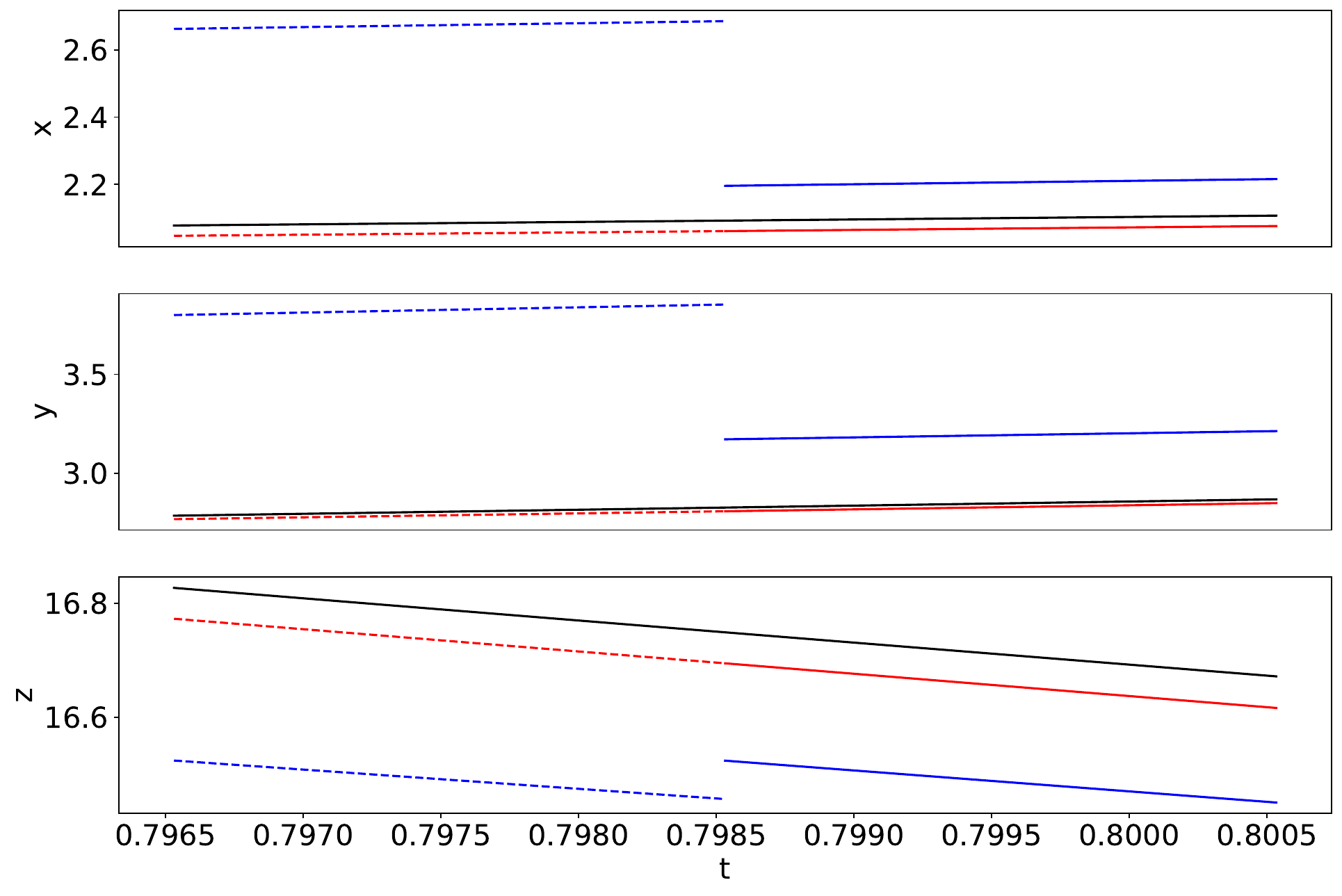}
    \caption{Zoomed in region of the computed trajectory where one time step ends and the next begins. }
    \label{fig:lorenz_cont}
\end{subfigure}
    \caption{Numerical results for the Lorenz--1963 system. \textit{Left:} Trajectories of the Lorenz--1963 system for the sONets and hONets. \textit{Right:} Zoomed in region between time steps highlighting the discontinuous/continuous nature of the obtained global solution for both methods.}
    \label{fig:lorenz_compare}
\end{figure}

We compare the errors of the two DeepONets for $100$ randomly chosen initial conditions after one time step, on the interval $[0,0.2]$, and after 10 time steps, on the interval $[0,2]$. When we evaluate the time stepping with each method and compare it to the corresponding reference solution, we observe that some trajectories significantly differ at some points from the reference solution. This is due the chaotic nature of the Lorenz system. We found that the soft constraint starts to differ for $8\%$ of the trajectories while the hard constraint only differs for $1\%$ of the trajectories. Moreover, when a trajectory computed with the hard-constrained DeepONet deviates from the corresponding reference trajectory, then the trajectory computed with the soft-constrained DeepONet does the same. 
This is why for computing the error, we only choose trajectories that stay close to the reference solution, see Table~\ref{tab:lorenz_error}. 
In Fig.~\ref{fig:lorenz_compare_good}, we show a numerical simulation for one such trajectory. 
In Fig.~\ref{fig:lorenz_cont}, we demonstrate that the soft constraint does not lead to a continuous simulation while the hard constraint ensures that the simulation is indeed continuous. The discontinuities in the simulation of the soft constraint can also be seen in Fig.~\ref{fig:lorenz_compare_good} and they lead to a greater deviation from the reference solution.

\begin{table}[t]
\centering
\begin{tabular}{l||c||c}
  Lorenz--1963   & NRMSE (1 step)        & NRMSE (10 steps)     \\ \hline\hline
sONet &         $1.02\cdot10^{-2}$ &           $6.83\cdot10^{-2}$       \\ \hline 
hONet & $\mathbf{1.31\cdot10^{-3}}$ & $\mathbf{3.80\cdot10^{-2}}$\\ \hline \hline 
Error reduction & 87\,\% & 44\,\%
\end{tabular}
\caption{Normalized root mean square error (NRMSE) for the soft and hard constraint simulation of the Lorenz system~\eqref{eq:Lorenz1963system}, and reduction in error obtained by using the hard constraint over the soft constraint, after $1$ and $10$ time steps corresponding to the time intervals $[0,0.2]$ and $[0,2]$. }\label{tab:lorenz_error}
\end{table}

\subsection{One-dimensional Poisson equation}\label{sec:PoissonEq}

The family of boundary value problem for one-dimensional Poisson equations of the form $u''(x) = f(x)$ on the interval $[-1,1]$ 
with the family of prescribed reference solutions 
\[
\uref(x;a,b)= \sin(3x) + a x + b
\]
is given by
\begin{gather}\label{eq:1DPoissonBVP}
\frac{\p^2 u}{\p x^2} = -9\sin(3x),\quad x\in [-1,1],\quad
u(-1)=\uref(-1;a,b),\quad  
u(1)=\uref(1;a,b).
\end{gather}
The boundary conditions are parameterized by the pair of constant boundary values, 
\[\boldsymbol\Phi=\mathbf g=(g_{\rm l},g_{\rm r})=\big(\uref(-1;a,b),\uref( 1;a,b)\big).\]
We train the networks under consideration on $10\,000$ values of the triple $(a,b,x)$, 
where each of values of the parameter~$a$ and~$b$ and the collocation points~$x$ is sampled uniformly in $[-1, 1]$.

An obvious ansatz for hard constraints for the problem~\eqref{eq:1DPoissonBVP} is 
\begin{gather}\label{eq:1DPoissonHONet1-3Ansatz}
u(x)\approx\mathcal G^\theta(\mathbf g)(x)=g_{\rm l}F_{\rm l}(x)+g_{\rm r}F_{\rm r}(x) + F_{\rm nn}(x)\mathcal G^\theta_{\rm t}(\mathbf g)(x),
\end{gather}
where the functions~$F_{\rm l}$, $F_{\rm l}$ and~$F_{\rm nn}$ satisfy the conditions 
$F_{\rm l}(-1)=F_{\rm r}(1)=1$, $F_{\rm l}(1)=F_{\rm r}(-1)=F_{\rm nn}(\pm1)=0$ and $F_{\rm nn}(x)\ne0$, $x\in(-1,1)$.
We test the following choices for these functions:
\begin{gather}\label{eq:1DPoissonHONet1Coeffs}
F_{\rm l}(x)=\frac{1-x}2,\quad F_{\rm r}(x)=\frac{1+x}2,\quad F_{\rm nn}(x)=1-x^2,
\\ \label{eq:1DPoissonHONet2Coeffs}
F_{\rm l}(x)=\frac{2+x}4(1-x)^2,\quad F_{\rm r}(x)=\frac{2-x}4(1+x)^2,\quad F_{\rm nn}(x)=1-x^2,
\\ \label{eq:1DPoissonHONet3Coeffs}
F_{\rm l}(x)=\frac{(1-x)^2}4,\quad F_{\rm r}(x)=\frac{(1+x)^2}4,\quad F_{\rm nn}(x)=1-x^2.
\end{gather}

We also incorporate the more sophisticated hard constraint that takes into account the expression for the second derivative of~$u$ 
in view of the equation to be solved,
\begin{gather}\label{eq:1DPoissonHardConstraintWith2ndOrderDers}
\begin{split}
u(x)\approx\mathcal G^\theta(\mathbf{g})(x)
={}&g_{\rm l}F_{\rm l0}(x)+g_{\rm r}F_{\rm r0}(x)
+ F_{\rm l1}(x)\mathcal G^\theta_{u'_{\rm l}\rm t}(\mathbf g) + F_{\rm r1}(x)\mathcal G^\theta_{u'_{\rm r}\rm t}(\mathbf g)\\
&{}- 9\sin(-3)F_{\rm l2}(x) - 9\sin(3)F_{\rm r2}(x)
+ F_{\rm nn}(x)\mathcal G^\theta_{\rm t}(\mathbf g)(x),
\end{split}
\end{gather}
where $\mathcal G^\theta_{u'_{\rm l}\rm t}(\mathbf g)$ and $\mathcal G^\theta_{u'_{\rm r}\rm t}(\mathbf g)$ 
are auxiliary trainable subnetworks that approximate the values of the first derivative of~$u$ 
at the left and right boundary points~$-1$ and~$1$, respectively, 
and the coefficients $F_{{\rm l}j}$, $F_{{\rm r}j}$ and~$F_{\rm nn}$ are functions of~$x$ satisfying the conditions
\begin{gather}\label{eq:1DPoissonHardConstraintWith2ndOrderDersCoeffConditions}
\begin{split}&
\frac{{\rm d}^jF_{{\rm l}j}}{{\rm d}t^j}(-1)=\frac{{\rm d}^jF_{{\rm r}j}}{{\rm d}t^j}( 1)=1,\quad  
\frac{{\rm d}^jF_{{\rm l}j}}{{\rm d}t^j}( 1)=\frac{{\rm d}^jF_{{\rm r}j}}{{\rm d}t^j}(-1)=0,\quad  
\\& 
\frac{{\rm d}^kF_{{\rm l}j}}{{\rm d}t^k}(\pm1)=\frac{{\rm d}^kF_{{\rm r}j}}{{\rm d}t^k}(\pm1)=\frac{{\rm d}^jF_{\rm nn}}{{\rm d}t^j}(\pm1)=0,\quad  
j,k\in\{0,1,2\},\quad k\ne j,
\\&
\frac{{\rm d}^3 F_{\rm nn}}{{\rm d}t^3}(\pm1)\ne0,\quad 
F_{\rm nn}(t)\ne0,\ t\in(-1,1). 
\end{split}
\end{gather}
We choose the following coefficients:
\begin{gather}\label{eq:1DPoissonHardConstraintWith2ndOrderDersCoeff}
\begin{split}&
F_{\rm l0}=\frac{8+9x+3x^2}{16}(1-x)^3,\ \ F_{\rm l1}=\frac{5+8x+3x^2}{16}(1-x)^3,\ \ F_{\rm l2}=\frac{(1+x)^2}{16}(1-x)^3,\\&
F_{\rm r0}=\frac{8-9x+3x^2}{16}(1+x)^3,\ \ F_{\rm r1}=\frac{5-8x+3x^2}{16}(1+x)^3,\ \ F_{\rm r2}=\frac{(1-x)^2}{16}(1+x)^3,\\&
F_{\rm nn}=(1-x^2)^3.
\end{split}
\end{gather}

\begin{figure}[!ht]
    \centering
    \includegraphics[width=0.5\textwidth]{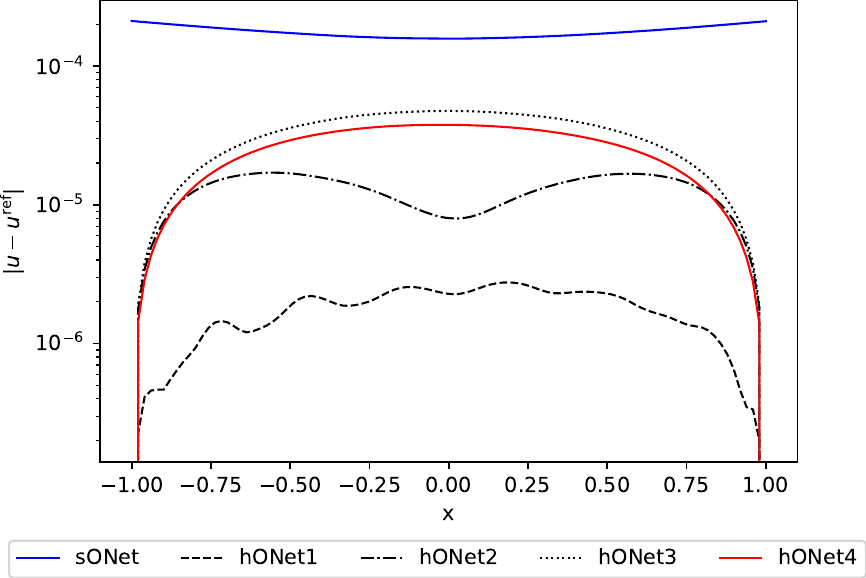}
    \caption{Absolute error of the hard- and soft-constrained solutions for the one-dimensional Poisson equation 
             averaged over 100 random samples of $(a,b)$ and 10 runs.}
    \label{fig:poisson_compare}
\end{figure}

In total, we train 
\begin{itemize}\itemsep=0ex
\item
one DeepONet with the soft constraint (referred to as sONet),
\item
three DeepONets with the hard constraints  
that are defined by the general ansatz~\eqref{eq:1DPoissonHONet1-3Ansatz}
with the coefficients~\eqref{eq:1DPoissonHONet1Coeffs},  
\eqref{eq:1DPoissonHONet2Coeffs} and~\eqref{eq:1DPoissonHONet3Coeffs}, 
which we refer to as hONet1, hONet2 and hONet3, respectively, and
\item
one DeepONet, hONet4, with the more sophisticated hard constraint defined 
by the general ansatz~\eqref{eq:1DPoissonHardConstraintWith2ndOrderDersCoeffConditions} 
with the coefficients~\eqref{eq:1DPoissonHardConstraintWith2ndOrderDersCoeff}.
\end{itemize}
We train 10 instances of each of the above networks with the soft and hard constraints, evaluate the trained networks on $100$ randomly chosen values of the pair $(a,b)$ and then average the obtained NRMSEs and absolute errors over all the evaluations, see Table~\ref{tab:poisson_error} and Fig.~\ref{fig:poisson_compare}. 
In Fig.~\ref{fig:poisson_compare}, we observe that the largest difference between the reference solutions and the sONet is at the boundary, in contrast to all the hONets, which exactly satisfy the boundary conditions. 
All the hONets essentially outperform the sONet. 
The exceptional results of the rather simple hONet1 is accidental,
owing to the fact that the hard constraint~\eqref{eq:1DPoissonHONet1-3Ansatz}, \eqref{eq:1DPoissonHONet1Coeffs} 
eliminates the parameters~$a$ and~$b$ from the training process. 
In other words, the trainable part~$\mathcal G^\theta_{\rm t}(\mathbf g)(x)$ of the network~$\mathcal G(\mathbf g)(x)$ 
does not depend on these parameters, $\mathcal G^\theta_{\rm t}(\mathbf g)(x)=\mathcal G^\theta_{\rm t}(x)$.
The numbers of layers and of units per layers in the auxiliary trainable subnetworks 
\smash{$\mathcal G^\theta_{u'_{\rm l}\rm t}(\mathbf g)$} and \smash{$\mathcal G^\theta_{u'_{\rm r}\rm t}(\mathbf g)$} 
in hONet4 are the same as those in the trunk and branch subnetworks, 
5 layers and 60 units per layers supplemented with the layer applying the linear activation.
Decreasing the number and the size of layers in these subnetworks
significantly worsen the accuracy of numerical solutions.
As expected, hONet4 better approximates the reference solutions than hONet2 and hONet3 close to the boundaries, 
and the performance of hONet2 is overall better than the performance of hONet3. 

\begin{table}[t]
\centering
\begin{tabular}{l||c}
\hfil 1D Poisson   & NRMSE \\ \hline\hline
sONet  & $2.3\cdot10^{-4}$ \\ \hline
hONet1 & $\boldsymbol{2.3\cdot10^{-6}}$ \\ \hline
hONet2 & $1.8\cdot10^{-5}$ \\ \hline
hONet3 & $3.9\cdot10^{-5}$ \\ \hline
hONet4 & $3.5\cdot10^{-5}$ \\ \hline\hline
Error reduction 1 & {\bf 99\,\%} \\ \hline
Error reduction 2 & 92\,\% \\ \hline
Error reduction 3 & 83\,\% \\ \hline
Error reduction 4 & 85\,\% 
\end{tabular}
\vspace{1ex}
\caption{Normalized root mean square error (NRMSE) 
for the soft and hard constraint simulations of the one-dimensional Poisson equation that is averaged over 10 training runs and 100 random samples of $(a,b)$,
and reduction in error obtained by using the hard constraints over the soft constraint.}\label{tab:poisson_error}
\end{table}

\noprint{

}

\subsection{One-dimensional linear wave equation}\label{sec:WaveEq}

To demonstrate that the hard-constraint approach works efficiently for non-evolution equations as well, 
we apply it to the family of initial--boundary value problems for the one-dimensional linear wave equations given by
\begin{gather}\label{eq:(1+1)DWaveEqBVP}
\begin{split}&
\frac{\p^2 u}{\p t^2}-\frac{\p^2 u}{\p x^2} =0,\quad (t,x)\in (0,t_{\rm f})\times(0, 1),\\&
u(t,0)=u(t,1)=0, \quad t\in [0,t_{\rm f}], \\&
u(0,x)=u_0(x;\mathbf{a}),\quad \frac{\p u}{\p t}(0,x) =0,\quad x\in [0,1], 
\end{split}
\end{gather}
cf.\ \cite[Section~4.3]{wang23a}.

The vanishing second initial condition and the vanishing Dirichlet boundary conditions essential simplify 
ansatzes for hard constraints for the problem~\eqref{eq:1DPoissonBVP}. 
We use the ansatz 
\[
u(x)\approx\mathcal G^\theta(\mathbf{a};\mathbf{\Phi})(x)
=u_0(x;\mathbf{a})F_{\rm i0}(t)+F_{\rm nn}(t,x)\mathcal G^\theta_{\rm t}(\mathbf{a};\mathbf{\Phi})(t,x),
\]
where the functions~$F_{\rm i0}(t)$ and~$F_{\rm nn}$ should satisfy the conditions 
\begin{gather*}
F_{\rm i0}(0)=1,\quad 
\frac{{\rm d}F_{\rm i0}}{{\rm d}t}(0)=0,\quad  
F_{\rm nn}(0,x)=\frac{\p F_{\rm nn}}{\p t}(0,x)=0,\ x\in[0,1],\\
F_{\rm nn}(t,0)=F_{\rm nn}(t,1)=0,\ t\in[0,t_{\rm f}],\quad 
F_{\rm nn}(t,x)\ne0,\ t\in(0,t_{\rm f}],\ x\in(0,1).
\end{gather*}
We test the following choices for these functions:
\begin{gather}\label{eq:(1+1)DWaveEqHONet1Coeffs}
F_{\rm i0}(t)=(1-t)^2(2t+1),\quad F_{\rm nn}(t,x)=t^2(3-2t)x(x-1),
\\ \label{eq:(1+1)DWaveEqHONet2Coeffs}
F_{\rm i0}(t)=\cosh^{-2}(4t),\quad F_{\rm nn}(t,x)=\tanh^2(4t)\sin(\pi x).
\end{gather}

For training, we use the parameterized set of initial values in the form of a truncated Fourier series, 
\[
u_0(x;\mathbf{a})=\sum_{n=1}^3 a_n \sin(n\pi x),\quad \text{where}\quad \mathbf{a}=(a_1,a_2,a_3)\in[-1,1]^3.
\]
and sample $10^6$ values for $(t,x,a_1,a_2,a_3)$, 
where the tuple $(t,x)$ is taken using Latin hypercube sampling in $[0,1]^2$
and each of~$a_n$, $n=1,2,3$, is sampled uniformly in $[-1,1]$.
The initial conditions are parameterized by their values at $N_{\rm s}=100$ evenly spaced sensor points $x_{\rm s}^j$, $j=1,\dots,100$, in $[0,1]$, and hence 
$\boldsymbol\Phi=\big(u_0(x_{\rm s}^1;\mathbf{a}),\dots,u_0(x_{\rm s}^{100};\mathbf{a})\big)$. 
We train 10 instances of each of the above DeepONets on the time interval $[0,1]$. 
Then we evaluate the trained networks for $100$ randomly chosen values of the triple $(a_1,a_2,a_3)$ 
and compare them after 1, 10 and 100 time steps of length~1. 
Note that due to the specific statement of the initial boundary value problem~\eqref{eq:(1+1)DWaveEqBVP}, 
namely, the zero initial derivative with respect to~$t$, the time stepping is possible only with steps of length~1, 
and for each step we set the initial derivative with respect to~$t$ to be equal to zero. 
We repeat the computation 10 times for each of the DeepONets under consideration. 
For error evaluation, we use the explicit solutions of the problem~\eqref{eq:(1+1)DWaveEqBVP} for the above initial values, 
which are respectively given by 
\begin{equation}
\uref(t,x)=\sum_{n=1}^3 a_n\cos(n\pi t)\sin(n\pi x),
\end{equation}
and average the computed error over the evaluation samples of $(a_1,a_2,a_3)$ and the runs.

\begin{figure}[!th]
\centering
\includegraphics[width=0.31\textwidth]{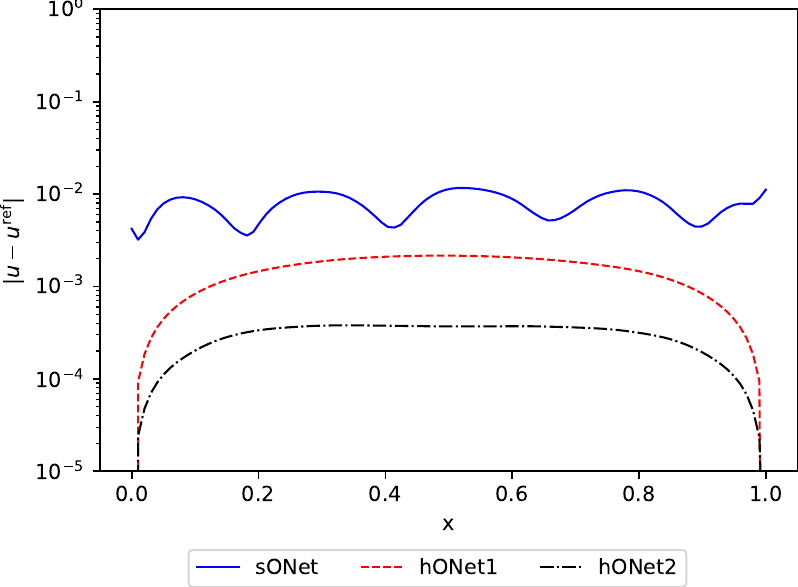}\quad
\includegraphics[width=0.31\textwidth]{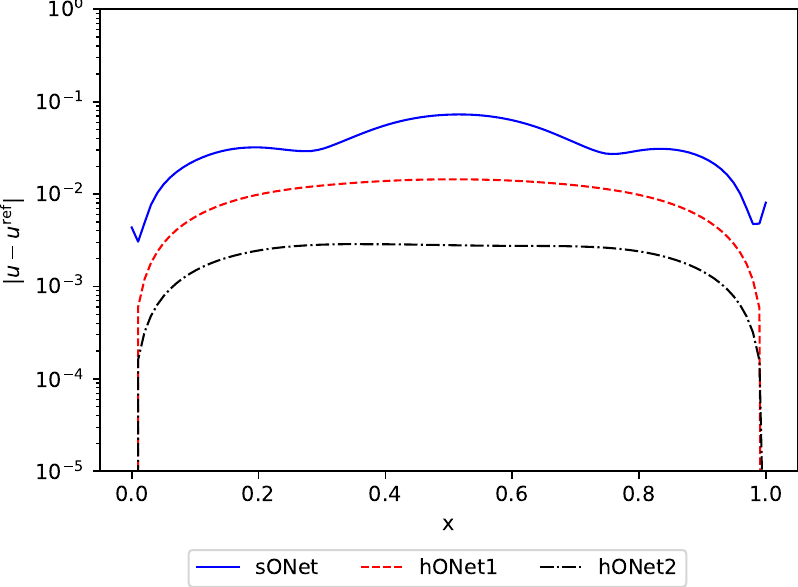}\quad
\includegraphics[width=0.31\textwidth]{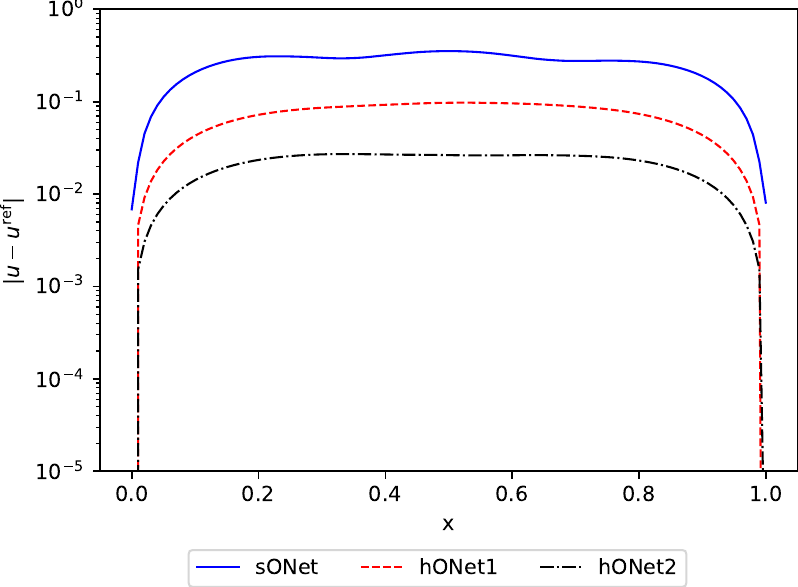}
\\[2ex]
\centering\tabcolsep=0ex
\begin{tabular}{lll}
\includegraphics[scale=0.43]{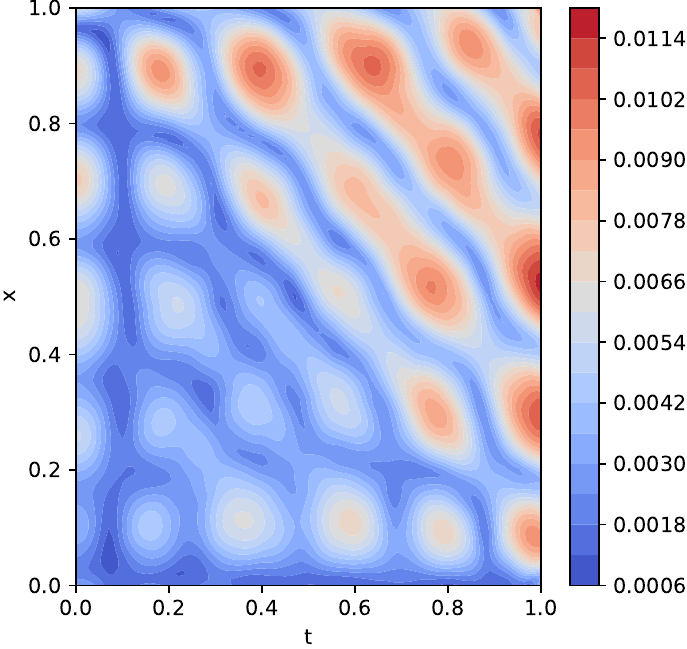}\hspace*{1em}&
\includegraphics[scale=0.43]{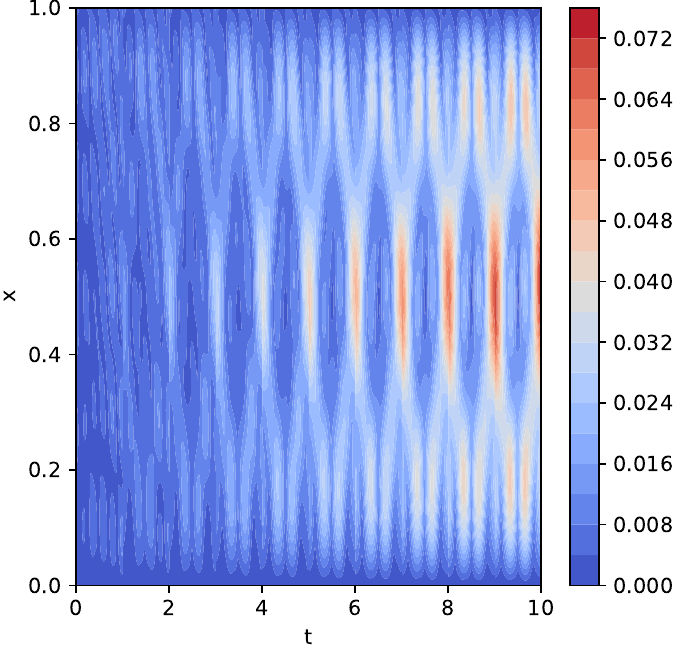}\hspace*{1em}&
\includegraphics[scale=0.43]{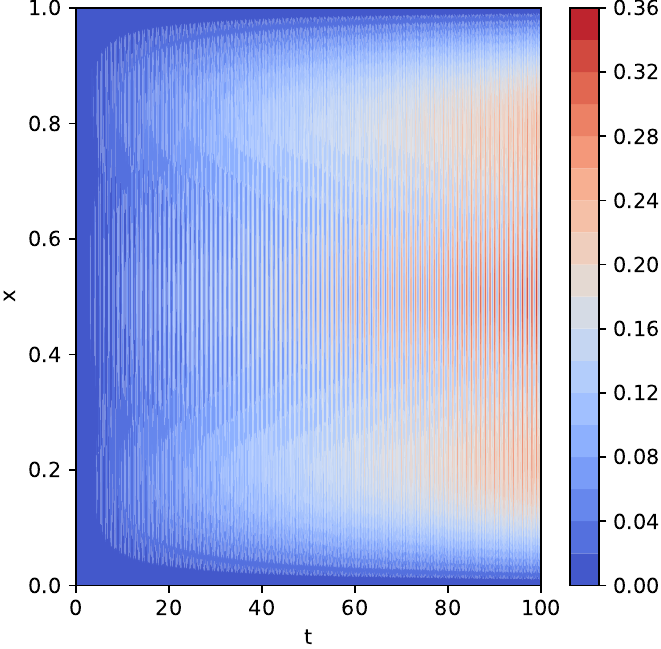}
\\[1ex]
\includegraphics[scale=0.43]{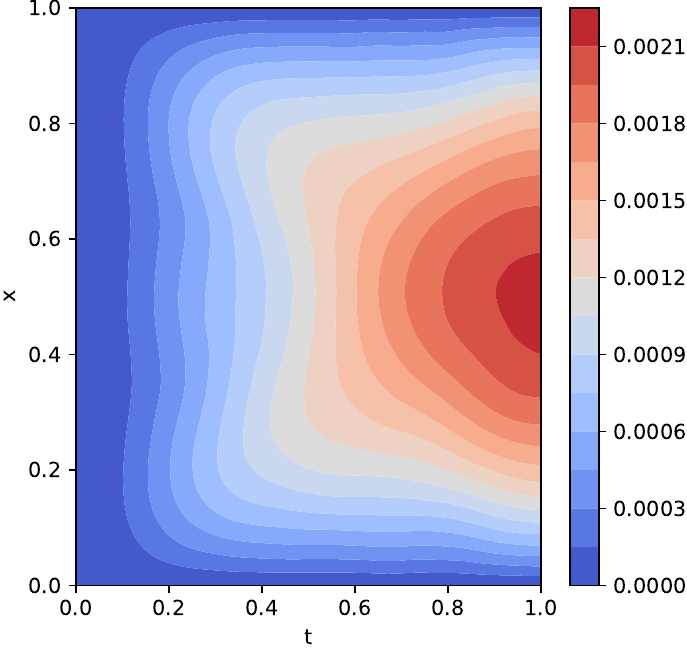}&
\includegraphics[scale=0.43]{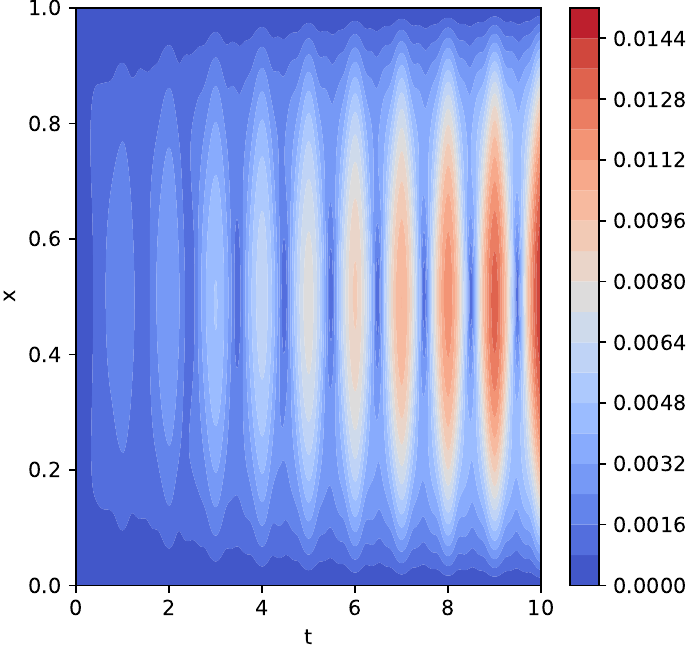}&
\includegraphics[scale=0.43]{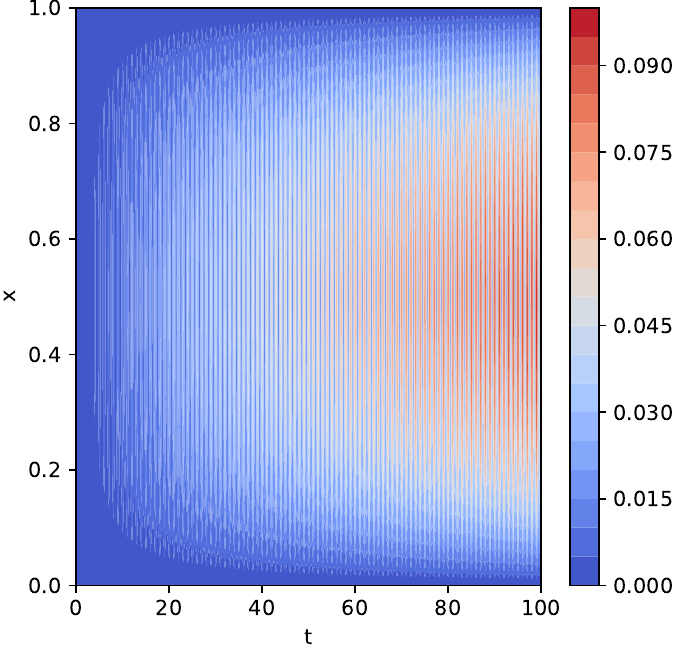}
\\[1ex]
\includegraphics[scale=0.43]{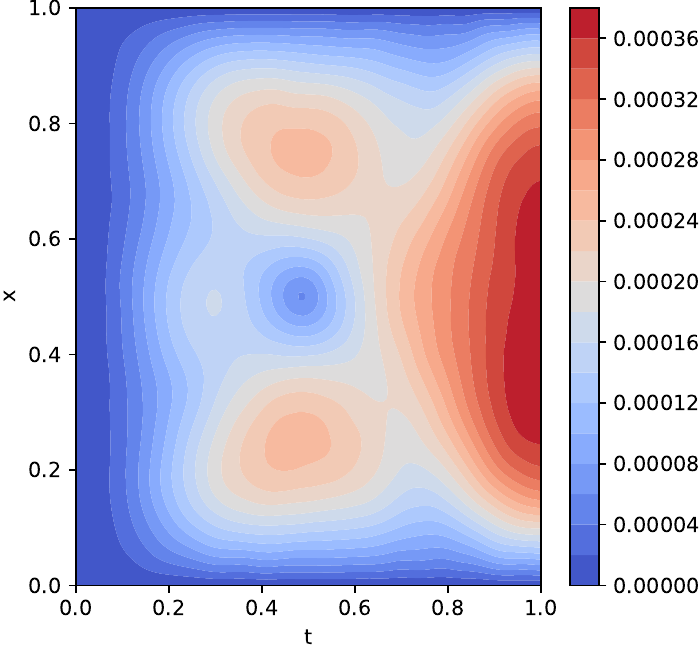}&
\includegraphics[scale=0.43]{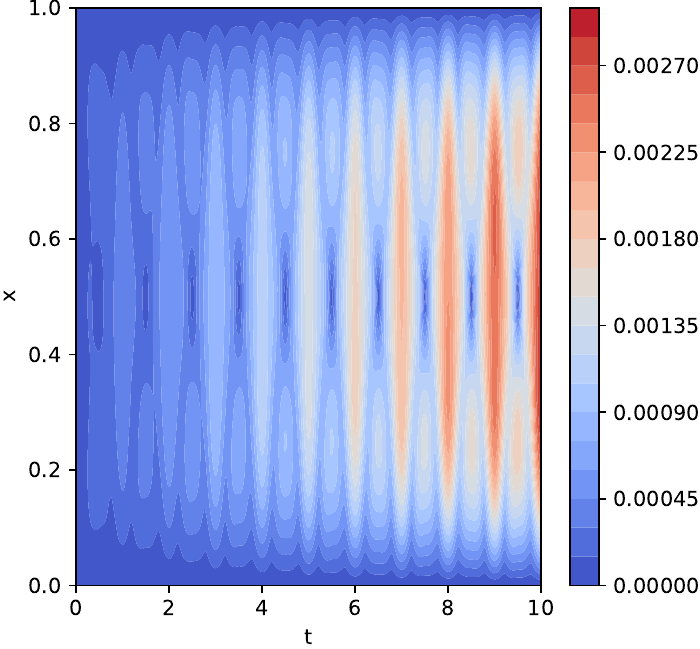}&
\includegraphics[scale=0.43]{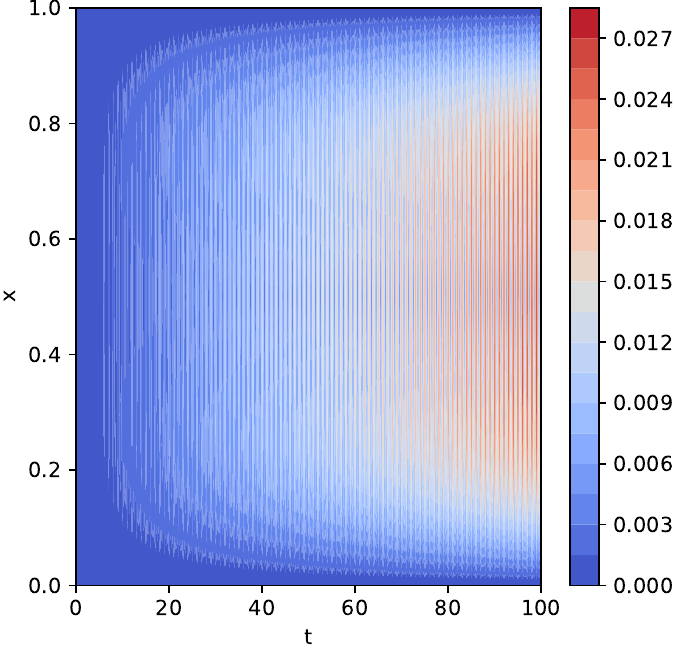}
\end{tabular}
\caption{Simulation error for the wave equation.
\textit{Left to right:} Evaluation of averaged absolute errors for $t$ in $[0,1]$, $[0,10]$ and $[0,100]$.
\textit{Top to bottom:} 
Comparison of the three considered DeepONets at the end of the time interval 
and averaged absolute errors for the sONet, the hONet1 and the hONet2, respectively.}
\label{fig:wave_compare}
\end{figure}

In Fig.~\ref{fig:wave_compare}, it is evident that the wave equation is more accurately represented using the hONets in comparison to the sONet, particularly noticeable during longer integration intervals. 
The sONet, which only approximates the initial and boundary conditions, has deviations as seen in the error plots, with errors at $t=0$ and at $x=0$ and $x=1$. 
Conversely, the hONets precisely satisfy the initial and boundary conditions.

The overall errors are assessed using the NRMSE, as reported in Table~\ref{tab:wave_error}. 
Again, both hONets essentially outperform the sONet for the single interval $[0,1]$ 
and in the course of time stepping, whereas the more elaborate constraint of hONet2 with trigomonetric and hyperbolic functions 
gives more accurate results than the simpler constraint of hONet1 with polynomial coefficients.

\begin{table}[!ht]
\centering
\begin{tabular}{l||c||c||c}
Wave  & NRMSE (1 step)    & NRMSE (10 steps) & NRMSE (100 steps)   \\ \hline\hline
sONet  & $1.2\cdot10^{-2}$& $4.6\cdot10^{-2}$& $3.1\cdot10^{-1}$\ \\\hline
hONet1 & $2.8\cdot10^{-3}$& $1.2\cdot10^{-2}$& $9.2\cdot10^{-2}$\ \\\hline
hONet2 & $5.2\cdot10^{-4}$& $2.7\cdot10^{-3}$& $2.5\cdot10^{-2}$\ \\\hline\hline
Error reduction 1 & 77\% & 73\% & 70\% \\ \hline
Error reduction 2 & 96\% & 94\% & 92\% 
\end{tabular}
\vspace{1ex}
\caption{Normalized root mean square error (NRMSE) for the soft-constraint and the hard-constraint simulations of the (1+1)-dimensional linear wave equation compared with the reference solutions that is averaged over 10 training runs and 100 random samples of $(a_1,a_2,a_3)$, and reduction in error obtained by using the hard constraints over the soft constraint, measured for the intervals $[0,1]$, $[0,10]$ and $[0,100]$.}\label{tab:wave_error}
\end{table}

\subsection{Korteweg--de Vries equation}\label{sec:KdV}

As the last example, we consider a nonlinear partial differential equation with initial and Dirichlet boundary conditions. 
Specifically, inspired by~\cite[Section~4.5]{wang23a},
we study a two-parameter set of initial--boundary value problems $\mathcal K(a,c;t_0,t_{\rm f})$ 
for the Korteweg--de Vries equation given by
\begin{gather}\label{eq:KdVProblem}
\begin{split}
    \frac{\p u}{\p t} + \epsilon u\frac{\p u}{\p x} + \mu \frac{\p^3 u}{\p x^3}=0,\quad (t,x)\in (t_0,t_{\rm f})\times (0, 5),
    \\
    u(t_0,x)=u_0(x;a,c):=\uref(t_0,x;a,c),\quad x \in [0, 5],
    \\
    u(t,0)=g_{\rm l}(t;a,c):=\uref(t,0;a,c),\quad t \in [t_0,t_{\rm f}], 
    \\
    u(t,5)=g_{\rm r}(t;a,c):=\uref(t,5;a,c),\quad t\in [t_0,t_{\rm f}],
    \\
    \frac{\p u}{\p x}(t,5)=h(t;a,c):=\frac{\p\uref}{\p x}(t,5;a,c),\quad t\in [t_0,t_{\rm f}]
\end{split}
\end{gather}
with $\epsilon=0.12$ and $\mu=0.0008$. 
The initial and boundary conditions are chosen in such a way that the explicit solution for the problem $\mathcal K(a,c;t_0,t_{\rm f})$ is given by
\begin{gather*}
    \uref(t,x;a,c)=\frac{c}{2}\text{sech}^2\left( \frac{\sqrt{c}}{2}\Big(5(x-a)-\frac{ct}{10}\Big) \right).
\end{gather*}
By their definition, these conditions are consistent, 
$u_0(0;a,c)=g_{\rm l}(0;a,c)$, 
$u_0(5;a,c)=g_{\rm r}(0;a,c)$ and 
$(\p u_0/\p x)(5;a,c)=h(0;a,c)$.
Although the equation and the reference solutions in~\eqref{eq:KdVProblem} 
are chosen to be the same as in \cite[Section~4.5]{wang23a}, we in fact study different, more complicated problems, 
not relying on known solutions and properly imposing boundary conditions.

In the general form of used hard constraints, we fix the dependence of coefficients on~$x$,
\begin{gather*}
u(t,x)\approx\mathcal G^\theta(a,c;\mathbf{\Phi})(t,x)= 
T_{\rm i}(t_{\rm n})u_0(x;a,c) 
+\big(g_{\rm l}(t;a,c)-T_{\rm i}(t_{\rm n})g_{\rm l}(0;a,c)\big)\frac{(x-5)^2}{25}
\\\qquad{} 
+\big(g_{\rm r}(t;a,c)-T_{\rm i}(t_{\rm n})g_{\rm r}(t;a,c)\big)\frac{10x-x^2}{25}
+\big(h(t;a,c)-T_{\rm i}(t_{\rm n})h(0;a,c)\big)\frac{x(x-5)}5
\\\qquad{}
+T_{\rm nn}(t_{\rm n})\frac{x(x-5)^2}{18}\mathcal G^\theta_{\rm t}(a,c;\mathbf{\Phi})(t,x),  
\end{gather*}
where $t_{\rm n}:=(t-t_0)/(t_{\rm f}-t_0)$, 
and the functions~$T_{\rm i}$ and~$T_{\rm nn}$ satisfy at least the conditions
\begin{gather*}
T_{\rm i}(0)=1,\quad T_{\rm nn}(0)=0,\quad \frac{\p T_{\rm nn}}{\p t}(0)=1,\quad T_{\rm nn}(t)\ne0,\ \ t\in(0,1]. 
\end{gather*}
We test the following choices for the functions~$T_{\rm i}$ and~$T_{\rm nn}$:
\begin{gather*}
\mbox{hONet0}\colon\quad T_{\rm i}(t_{\rm n})=1,\quad  T_{\rm nn}(t_{\rm n})=t_{\rm n},\\
\mbox{hONet1}\colon\quad T_{\rm i}(t_{\rm n}):=1-3t_{\rm n}^{\,2}+2t_{\rm n}^{\,3}
,\quad T_{\rm nn}(t_{\rm n})=t_{\rm n},\\
\mbox{hONet2}\colon\quad T_{\rm i}(t_{\rm n}):=1-3t_{\rm n}^{\,2}+2t_{\rm n}^{\,3}
,\quad T_{\rm nn}(t_{\rm n})=t_{\rm n}+t_{\rm n}^2-t_{\rm n}^3,\\
\mbox{hONet3}\colon\quad T_{\rm i}(t_{\rm n}):=\cosh^{-2}(5t_{\rm n}),\quad T_{\rm nn}(t_{\rm n})=\tanh(4t_{\rm n}),
\end{gather*}
where on the left we write the notion of the corresponding hONet. 

The initial conditions are parameterized by their values at $N_{\rm s}=100$ evenly spaced sensor points $x_{\rm s}^j$, $j=1,\dots,100$, in $[0,5]$, 
and hence $\boldsymbol\Phi=\big(u_0(x_{\rm s}^1;a,c),\dots,u_0(x_{\rm s}^{100};a,c)\big)$. 
For training, we sample $200\,000$ different values of $(t,x,a,c)$
by Latin hypercube sampling of $(t,x)$ in $[0,t_{\rm f}]\times[0,5]$ and
by random uniform sampling of~$(a,c)$ in $[1,4]\times[10^{-5},2]$. 

All the model are trained to solve the problems of the form~\eqref{eq:KdVProblem}
for the single value of the initial time $t_0=0$ 
and two values for the final time, $t_{\rm f}=1$ and $t_{\rm f}=10$, 
$\mathcal K(a,c;0,1)$ and $\mathcal K(a,c;0,10)$.
Note that although we intend to use the trained DeepONet for time-stepping, 
it suffices to train it only for the initial time $t_0=0$ 
since time translations act as equivalence transformations on the set of the problems $\mathcal K(a,c;t_0,t_{\rm f})$ 
due to the following property of $\uref(t,x;a,c)$:
\[
\uref(t+\delta,x;a,c)=\uref(t,x;\tilde a,c),\quad\text{where}\quad\tilde a:=a+\frac{c\delta}{50}.
\]
In particular, $\uref(kt_{\rm f},x;a,c)=\uref(0,x;a+kt_{\rm f}c/50,c)$ and $2+k/5\in[1,4]$, $k=0,1,\dots$.
Hence each trained DeepONet for the set of the problems $\mathcal K(a,c;0,1)$ with $(a,c)\in[1,4]\times[10^{-5},2]$
can be used for 10 time steps in order to solve the problems $\mathcal K(a,c;0,10)$, 
where $c\in[10^{-5},2]$ and $a\in[1,4-0.18c]$.
Analogously, each trained DeepONet for the set of the problems $\mathcal K(a,c;0,10)$ with $(a,c)\in[1,4]\times[10^{-5},2]$
can be used for 10 time steps in order to solve the problems $\mathcal K(a,c;0,100)$, 
where $c\in[10^{-5},2]$ and $a\in[1,4-1.8c]$.

For the problems of the form~\eqref{eq:KdVProblem}, the training hyper-parameters needed an additional tuning 
for the training results to be reasonable, especially for sONets. 
This is why we used training settings 
that differ from those for the other specific problems considered in the present paper 
and, moreover, are in part different for the hONets and the sONet. 
We still utilize the Adam optimizer and preserve its parameters $\beta_1=0.95$ and $\beta_2=0.99$ 
and the exponentially decaying learning rate with the decay rate of 0.95 
but, at the same time, select the initial learning rate of $6\cdot10^{-3}$ and $2\cdot10^{-3}$ 
and the ``decay steps'' parameter of $140\times\mathsf B$ and $200\times\mathsf B$ 
for the hONets and the sONet, respectively. 
As in the introductory part of Section~\ref{sec:Results}, 
here $\mathsf B$ denotes the number of batches per epoch.
Each training was 10000 epochs. 
Moreover, for the sONet, we tune the weights~$\lambda_{\Delta}$, $\lambda_{\rm i}$ and~$\lambda_{\rm sb}$
of the equation loss, the initial value loss and the spatial boundary value loss 
in the composite loss function and set them to be equal to 1, 10 and 2, respectively. 

\begin{figure}[!ht]
\centering
\includegraphics[width=0.31\textwidth]{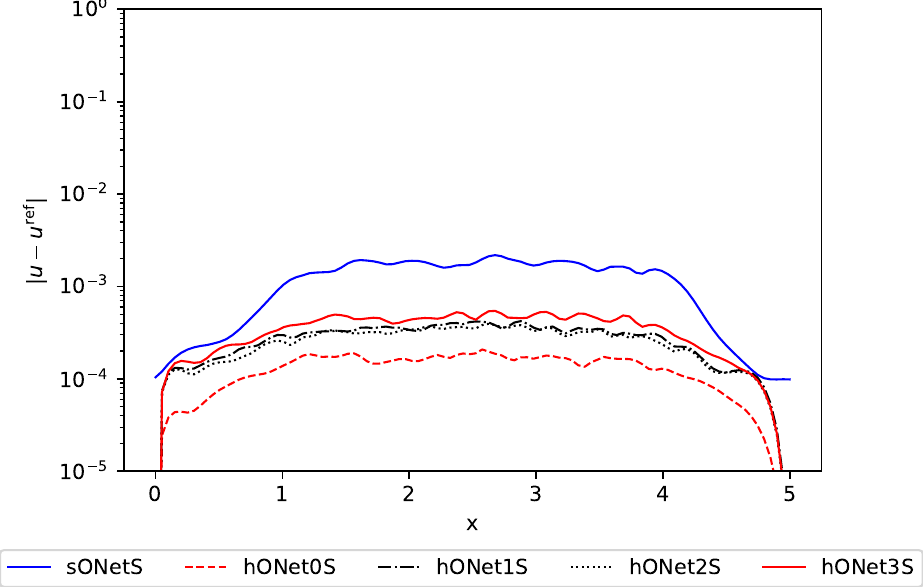}\quad
\includegraphics[width=0.31\textwidth]{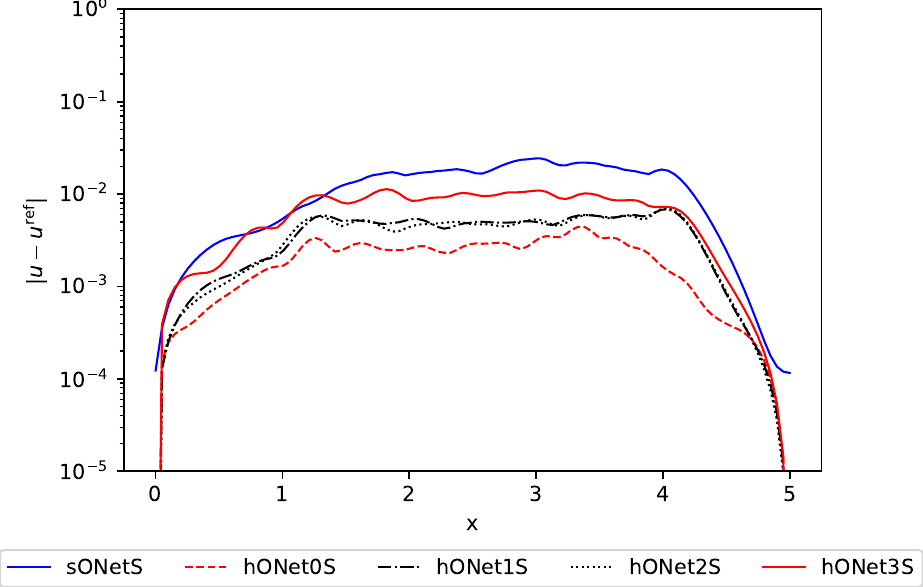}\quad
\includegraphics[width=0.31\textwidth]{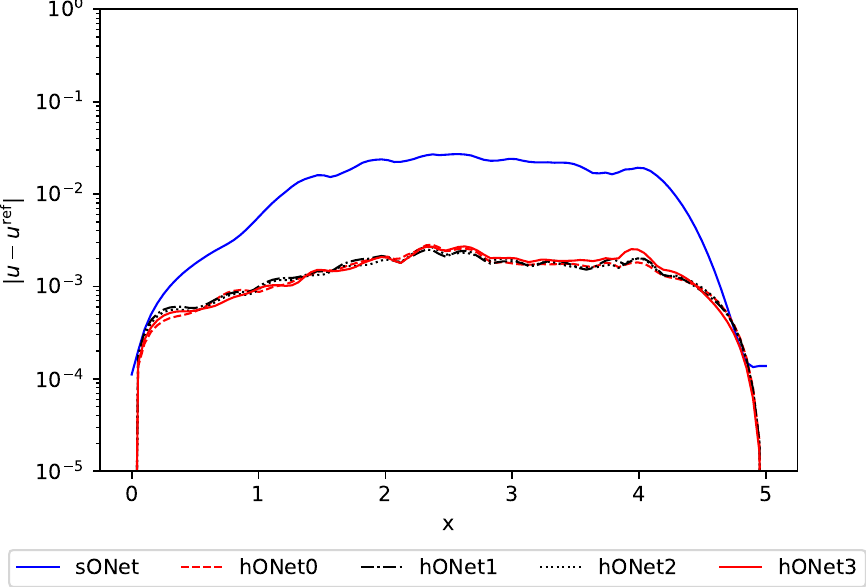}
\caption{Comparison of simulation error of the five considered DeepONets for the KdV equation 
at the end of the time interval.
\textit{Left to right:} Evaluation of averaged absolute errors for $t$ 
in the time intervals $[0,1]$, $[0,10]$ and $[0,10]$ run with 1, 10 and~1 steps, respectively.}
\label{fig:KdV_compareA}
\end{figure}

\begin{figure}
\centering\tabcolsep=0ex
\begin{tabular}{lll}
\includegraphics[scale=.41]{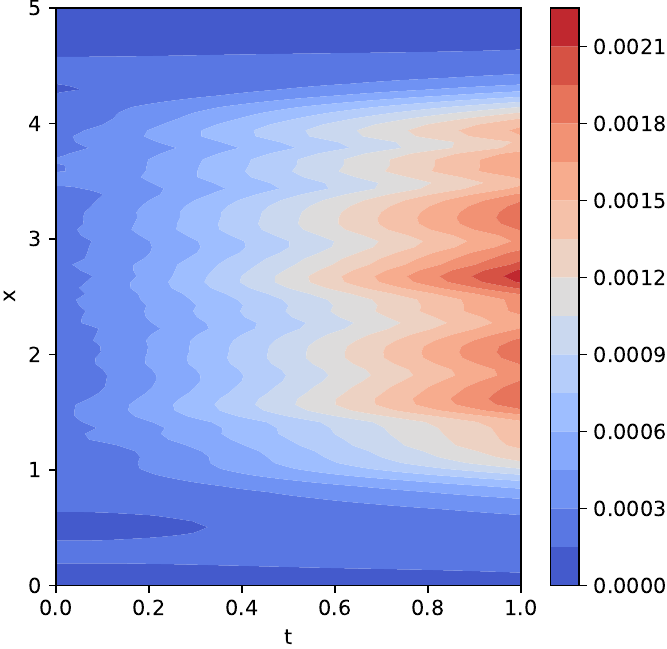} \hspace*{1em}&
\includegraphics[scale=.41]{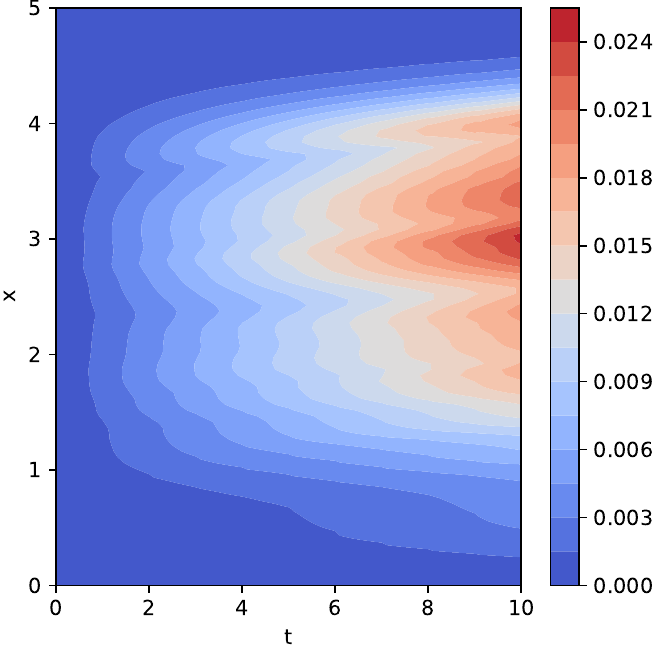}\hspace*{1em}&
\includegraphics[scale=.41]{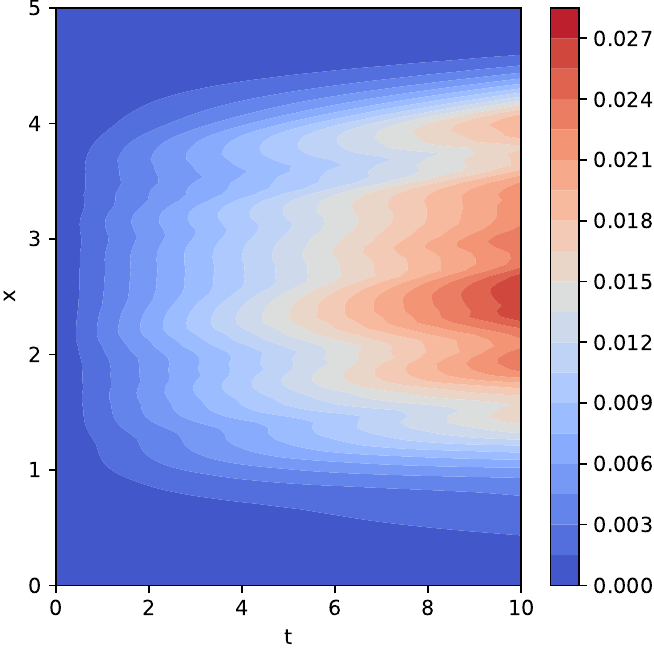}
\\
\includegraphics[scale=.41]{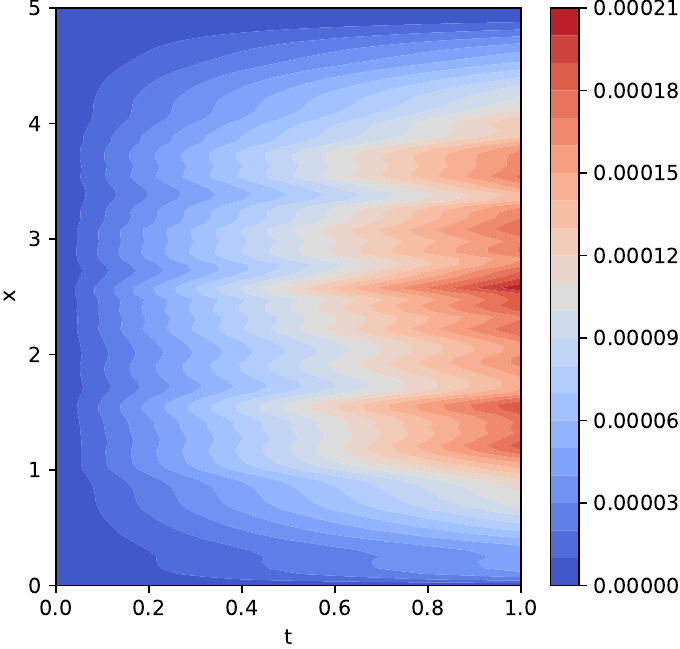} &
\includegraphics[scale=.41]{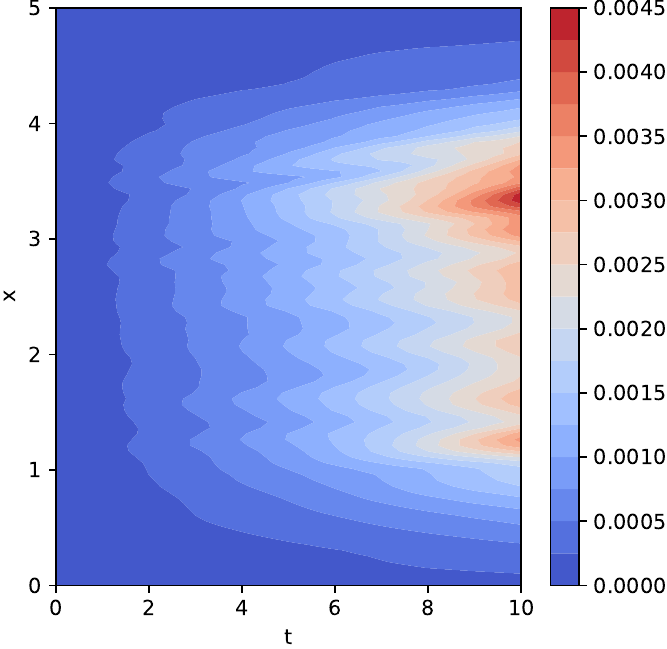}&
\includegraphics[scale=.41]{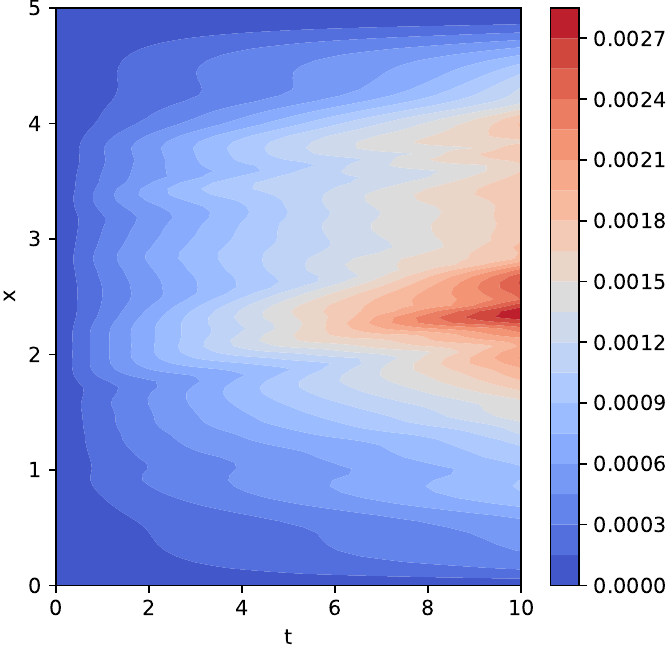}
\\
\includegraphics[scale=.41]{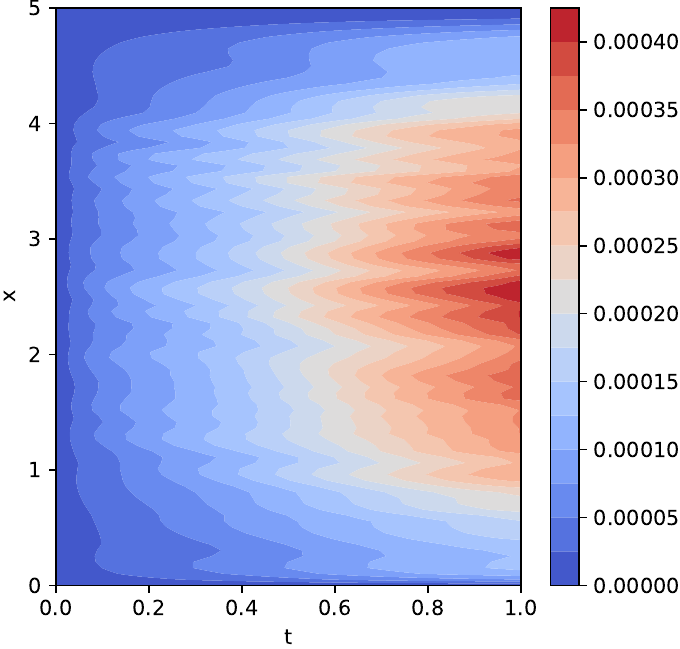} &
\includegraphics[scale=.41]{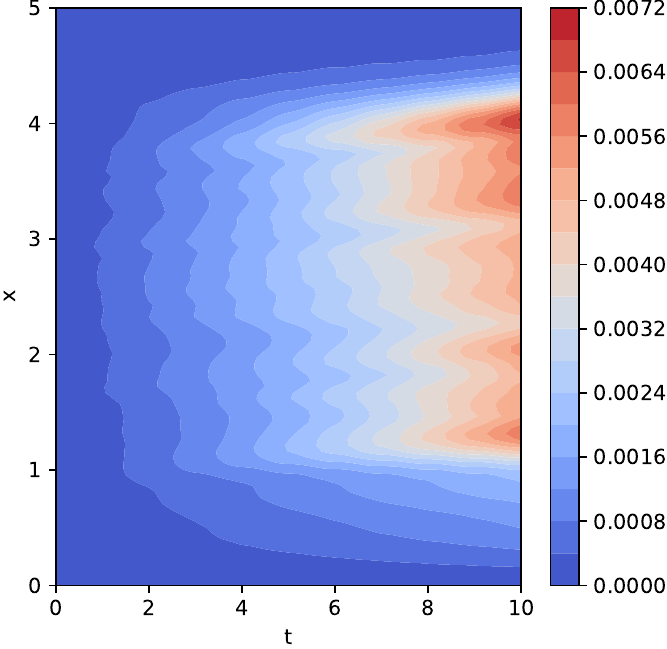}&
\includegraphics[scale=.41]{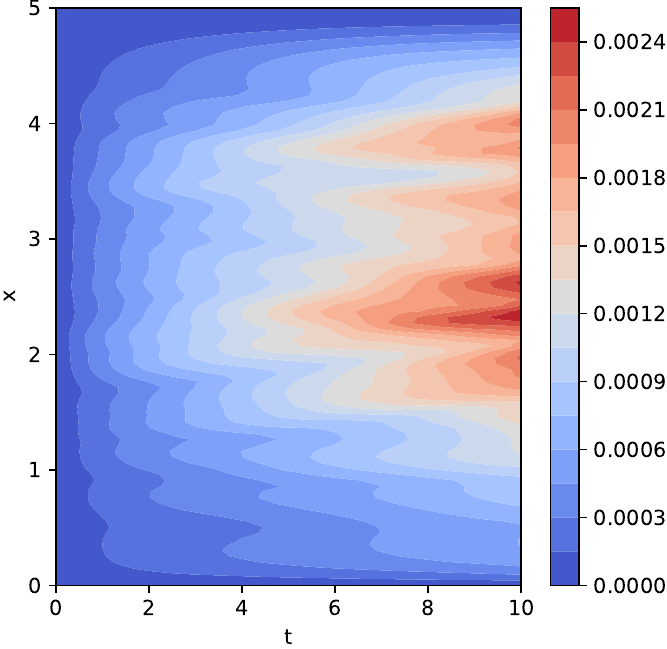}
\\
\includegraphics[scale=.41]{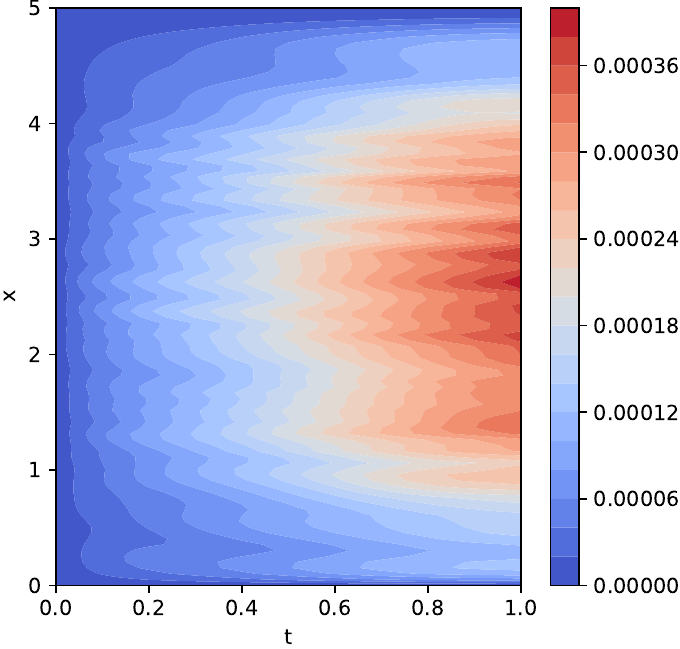} &
\includegraphics[scale=.41]{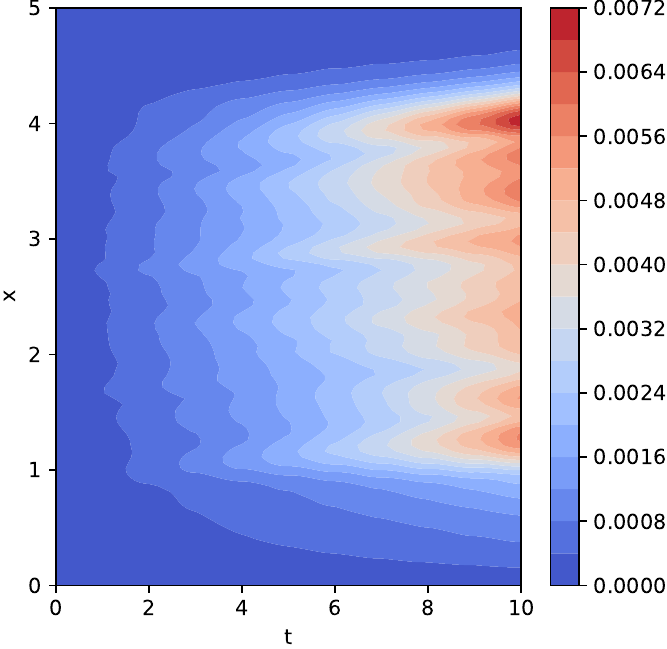}&
\includegraphics[scale=.41]{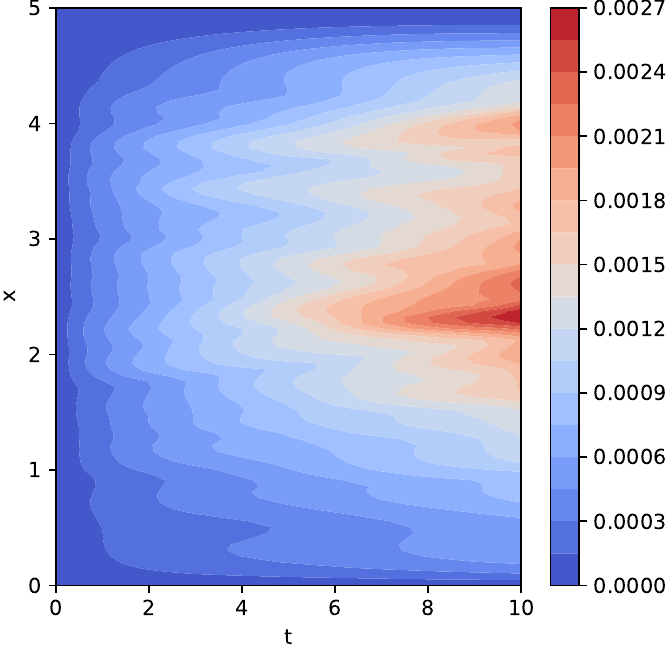}
\\
\includegraphics[scale=.41]{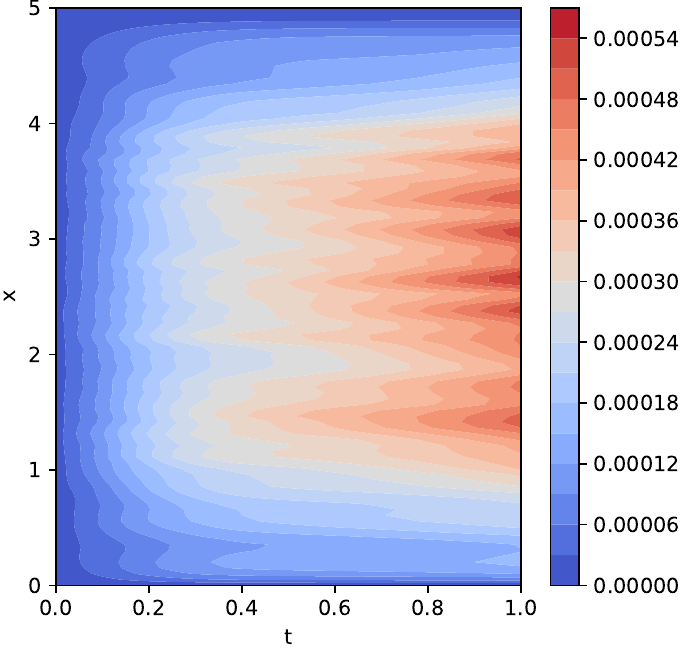} &
\includegraphics[scale=.41]{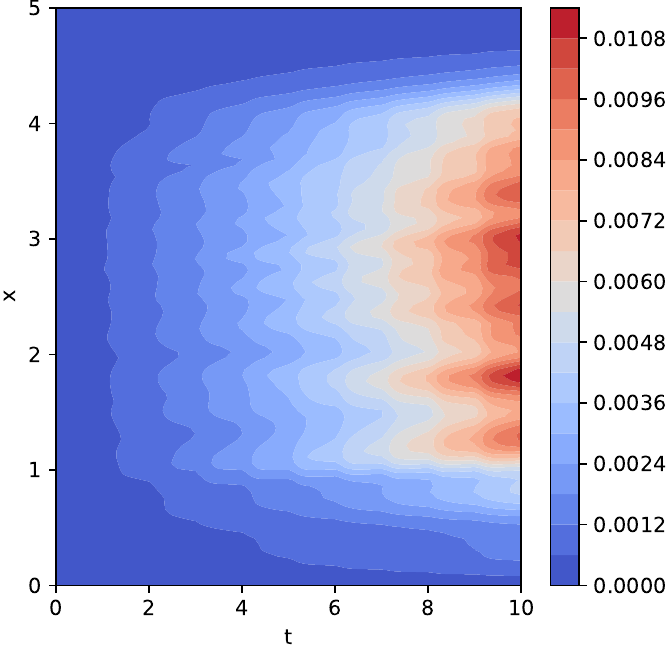}&
\includegraphics[scale=.41]{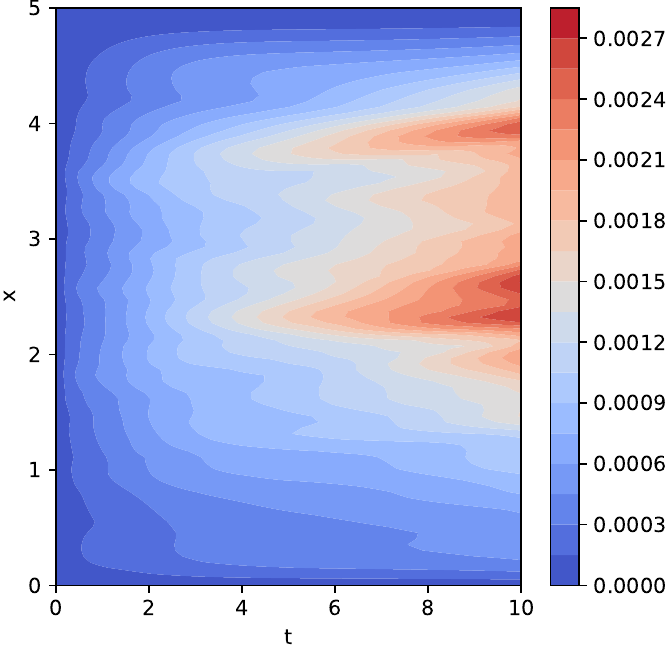}
\end{tabular}
\caption{Simulation error for the KdV equation.
\textit{Left to right:} Evaluation of averaged absolute errors for $t$ 
in the time intervals $[0,1]$, $[0,10]$ and $[0,10]$ run with 1, 10 and~1  steps, respectively.
\textit{Top to bottom:} 
Averaged absolute errors for the sONet and the hONet0, \dots, the hONet3, respectively.}
\label{fig:KdV_compareB}
\end{figure}

Similarly to the other considered problems, 
we train 10 instances of each of the above networks with the soft and hard constraints, evaluate the trained networks on $100$ randomly chosen values of the pair $(a,c)$ and then average the obtained NRMSEs and absolute errors over all the evaluations, 
see Table~\ref{tab:KdV_error} and Figs.~\ref{fig:KdV_compareA} and~\ref{fig:KdV_compareB}. 
Both the table and the figures evidently show that all the four hONets much more accurately approximate true solutions than the sONet does. 
The greatest difference in accuracy occurs when the training is conducted for the longer interval $[0,10]$.
It is unexpected that the best results on both time intervals $[0,1]$ and $[0,10]$ 
and in the course time stepping is achieved for the simplest hONet0, 
which is in contrast to the other considered problems. 
At the same time, the hONets are notably close in performance, when the basic time interval is $[0,10]$.

\begin{table}[!ht]
\centering
\begin{tabular}{l||c||c||c||c}
\hfil KdV  & NRMSE         & NRMSE           & NRMSE          & NRMSE 
\\          & (1 step by 1) & (10 steps by 1) & (1 step by 10) & (10 step by 10)
\\ \hline\hline
sONet  & $8.8\cdot10^{-3}$ & $8.2\cdot10^{-2}$ & $1.1\cdot10^{-1}$ & $2.4\cdot10^{ 0}$ \\ \hline
hONet0 & $\boldsymbol{9.9\cdot10^{-4}}$ & $\boldsymbol{1.1\cdot10^{-2}}$ & $\boldsymbol{9.1\cdot10^{-3}}$ & $\boldsymbol{3.5\cdot10^{-1}}$ \\ \hline
hONet1 & $1.9\cdot10^{-3}$ & $2.2\cdot10^{-2}$ & $9.1\cdot10^{-3}$ & $3.9\cdot10^{-1}$ \\ \hline
hONet2 & $1.8\cdot10^{-3}$ & $2.1\cdot10^{-2}$ & $9.0\cdot10^{-3}$ & $3.9\cdot10^{-1}$ \\ \hline
hONet3 & $2.7\cdot10^{-3}$ & $3.4\cdot10^{-2}$ & $1.0\cdot10^{-2}$ & $4.1\cdot10^{-1}$ \\ \hline\hline
Error reduction 1 & \bf 89\% & \bf 87\% & \bf 92\% & \bf 86\% \\ \hline
Error reduction 2 & 78\% & 74\% & 92\% & 84\% \\ \hline
Error reduction 3 & 79\% & 75\% & 92\% & 84\% \\ \hline
Error reduction 4 & 69\% & 59\% & 91\% & 83\% \\ \hline\hline
\end{tabular}
\vspace{1ex}
\caption{Normalized root mean square error (NRMSE) for the soft-constraint and the hard-constraint simulations of the KdV equation compared with the reference solutions that is averaged over 10 training runs and 100 random samples of $(a,c)$, and reduction in error obtained by using the hard constraints over the soft constraint, measured for the time intervals $[0,1]$, $[0,10]$, $[0,10]$ and $[0,100]$ run by 1, 10, 1 and~10 steps, respectively.}\label{tab:KdV_error}
\end{table}

\section{Conclusion}\label{sec:Conclusion}

This study pioneered the integration of a hard constraint initial and/or boundary condition into the architecture or loss function of a physics-informed DeepONet, thereby achieving exact representation of initial and boundary values. This hard constraint inclusion has several advantage. Firstly, it substantially simplifies the optimization problem that has to be solved to train physics-informed DeepONets. This is due to the loss function then reducing to the differential equation loss only, as all initial and/or boundary conditions are automatically satisfied by the model architecture. Secondly, when utilizing the DeepONet to perform time-stepping, a continuous global solution is guaranteed. This is in contrast to the traditional soft-constraint approach, where the solution will incur jumps when iterative time stepping is carried out. We have illustrated for several examples that the proposed hard-constraint method outperforms the traditional soft-constraint approach.

For the purpose of this paper, we have focused on the simplest form of the hard-constrained DeepONets. Further room for experimentation exists in using more sophisticated ansatzes that incorporate the initial and/or boundary data of a differential equation into the neural network exactly. This may be essential for ensuring not only continuity, but also differentiability of a time-stepped numerical solution using DeepONets. 
Note that more accurate results can be obtained by carefully selecting hard-constraint ansatzes and adjusting optimizer parameters
but in the present paper we pay the most attention to the relative comparison of the soft- and hard-constraint approaches.

To the best of our knowledge, DeepONets have rarely been applied to boundary value problems. The standard soft-constrained DeepONet needs to learn the boundary values in order to accurately learn the solution. In contrast, the hard-constraint does not suffer from this problem and thus achieves much better results.

We should also like to point out that while the hard-constrained DeepONets have generally given lower errors than their soft-constrained counterparts, there may be situations where the soft-constraint approach is more straightforward to implement. This concerns the formulation of problems on irregular domains, which may pose challenges for the hard-constrained version of DeepONets.

\section*{Acknowledgements}

This research was undertaken thanks to funding from the Canada Research Chairs program, the NSERC Discovery Grant program and the AARMS CRG \textit{Mathematical foundations and applications
of Scientific Machine Learning}. The research of RB is funded by the Deutsche Forschungsgemeinschaft (DFG, German Research Foundation) -- Project-ID 274762653 -- TRR 181.
The research of ROP was supported in part by the Ministry of Education, Youth and Sports of the Czech Republic (M\v SMT \v CR)
under RVO funding for I\v C47813059.
ROP also expresses his gratitude for the hospitality shown by the University of Vienna during his long staying at the university.

\footnotesize\setlength{\itemsep}{0ex}

\end{document}